\documentclass[11pt,letterpaper]{article}

\pdfoutput=1
\usepackage{epsfig}
\usepackage{graphicx}
\usepackage{amsmath}
\usepackage{amssymb}
\usepackage{geometry}

\usepackage{algorithm}
\usepackage{algorithmic}
\usepackage{caption}
\usepackage{subcaption}
\usepackage{mathtools}
\newcommand{\argmin}{\operatornamewithlimits{argmin}}

\newcommand{\bx}{\textbf{x}}
\newcommand{\by}{\textbf{y}}

\usepackage{color}
\usepackage{epstopdf}
\usepackage[square,numbers,sort&compress]{natbib}

\newcommand{\comments}[1]{}
\usepackage{mdwlist}

\makeatother

\usepackage[pagebackref=false,breaklinks=true,letterpaper=true,colorlinks,bookmarks=false]{hyperref}

\begin{document}

\title{Leveraging Long-Term Predictions and Online-Learning in Agent-based Multiple Person Tracking}

\author{Wenxi Liu$^{1}$ \hspace{0.2in} Antoni B. Chan$^{1}$ \hspace{0.2in}
Rynson W.H. Lau$^{1}$ \hspace{0.2in} Dinesh Manocha$^{2}$ \vspace{0.1in} \\
$^{1}$ Department of Computer Science, City University of Hong Kong, Hong Kong \\
$^{2}$ Department of Computer Science, University of North Carolina, Chapel Hill, U.S.A. \\
\vspace{-0.3in}}
\date{}
\maketitle

\begin{abstract}
We present a multiple-person tracking algorithm, based on combining particle filters and RVO, an agent-based crowd model that infers collision-free velocities so as to predict pedestrian's motion. In addition to position and velocity, our tracking algorithm can estimate the internal goals (desired destination or desired velocity) of the tracked pedestrian in an online manner, thus removing the need to specify this information beforehand. Furthermore, we leverage the longer-term predictions of RVO by deriving a higher-order particle filter, which aggregates multiple predictions from different prior time steps. This yields a tracker that can recover from short-term occlusions and  spurious noise in the appearance model. Experimental results show that our  tracking algorithm is suitable for predicting pedestrians' behaviors online without needing scene priors or hand-annotated goal information, and improves tracking in real-world crowded scenes under low frame rates.
\end{abstract}

\section{Introduction}

Pedestrian tracking from videos of indoor and outdoor scenes remains a challenging task despite recent advances. It is especially difficult in crowded real-world scenarios because of practical problems such as inconsistent illumination conditions, partial occlusions and dynamically changing visual appearances of the pedestrians. For applications such as video surveillance, the video sequence usually has relatively low resolution or low frame rate, and may thus fail to provide sufficient information for the appearance-based tracking methods to work effectively.

One method for improving the robustness of pedestrian tracking w.r.t. these confounding factors
is to use an accurate motion model to predict the next location of each pedestrian in the scene.
In the real-world, each individual person 
plans its path and determines its trajectory
based on both
its own internal goal (e.g., its final destination),
and the locations of any surrounding people (e.g., to avoid collisions, or to remain in a group).
Many recent methods \cite{pellegrini_youll_2009, yamaguchi_who_2011} have proposed online tracking using agent-based crowd simulators, which treat each person as a separate entity that independently plans its own path. 

Despite their successes, agent-based online trackers~\cite{pellegrini_youll_2009, yamaguchi_who_2011}  have several shortcomings.
First, they assume that the goal position or destination information of each pedestrian in the scene is known in advance, either hand annotated or estimated off-line using training data.
Second, these tracking methods only use single-step prediction, which is a short-term prediction of the pedestrian's next location  based on its current position. As a result, these trackers are not fully leveraging the capabilities of agent-based motion models to predict {\em longer-term} trajectories, e.g., steering around obstacles.
Third, existing motion models often overlook subtle, but crucial, factors in crowd interactions, including anticipatory motion and smooth trajectories~\cite{guy2012least}; this avoidance is \emph{reciprocal}, as pedestrians mutually adjust their movements to avoid anticipated collisions. One of our goals is to use the long-term predictive capacities of such motion models to improve pedestrian tracking.

In this paper, we address these shortcomings by proposing a novel multiple-person tracking algorithm, based on a higher-ordered particle filter and an agent-based motion model using reciprocal velocity obstacles (RVO).
RVO~\cite{van_den_berg_reciprocal_2011}, a multi-agent collision avoidance scheme, has
been used for multi-robot motion planning and generating trajectories of synthetic agents in dense crowds.
In this paper, we combine the RVO motion model with particle filter tracking,
resulting in a flexible multi-modal tracker that can estimate both the physical properties (position and velocity) and individual internal goals (desired destination or desired velocity).
This estimation is performed in an online manner without acquiring scene prior or hand-annotated goal information.
Leveraging the longer-term predictions of RVO, we derive a higher-ordered particle filter (HPF) based on a higher-order Markov assumption.  The HPF constructs a posterior that accumulates multiple predictions of the current location based on multiple prior time steps.
Because these predictions from the past ignore subsequent observations, the HPF tracker is
better able to recover from spurious failures of the appearance model and short-term occlusions that are common in multi-pedestrian tracking.

We evaluate the efficacy of our proposed tracking algorithm on low frame rate video of real scenes, and show that our approach can estimate internal goals online, and improve trajectory prediction and pedestrian tracking over prior methods.
In summary, the major contributions of this paper are three-fold:
%
\begin{itemize}
  \item We propose a novel tracking method that integrates agent-based crowd models and the particle filter, which can
  reason about pedestrians' motion and interaction and adaptively estimate their internal goals. (Sec.~\ref{sec:rvotrack})
  \item We propose a novel higher-order particle filter, based on a higher-order Markov model, to fully leverage the longer-term predictions of RVO. (Sec.~\ref{sec:HPF})
  \item We demonstrate that our RVO-based online tracking algorithm improves tracking in low frame rate crowd video, compared to other motion models, while not requiring scene priors or hand-annotated goal information. (Sec.~\ref{sec:results})
\end{itemize}

\section{Related works}
\label{sec:relatedwork}

Multi-object tracking has been an active research topic in recent years. It is a challenging problem due to the changing appearances caused by factors including pose, illumination, and occlusion. \cite{yilmaz_object_2006} gives a general introduction to object tracking.  Despite recent advances in tracking~\cite{MIL_PAMI_2011,hare2011struck}, appearance-based approaches struggle to handle large numbers of targets in crowded scenes.

\noindent{\bf Data association:}  Recent data association approaches for multi-object tracking~\cite{li2009learning,benfold2011stable,andriyenko_discrete-continuous_2012}
usually associate detected responses into longer tracks, which can be formulated as a global optimization problem,
e.g., by computing optimal network flow~\cite{zhang2008global,butt2013multi} or using linear programming~\cite{berclaz2009multiple,jiang2007linear}.

\noindent
{\bf Particle filters:} Particle filters (PF) are often used for online multi-object tracking~\cite{perez2002color,khan_mcmc-based_2004,vermaak2003maintaining,okuma2004boosted,breitenstein_robust_2009,bazzani2012decentralized,hess_discriminatively_2009}.
To handle the high-dimensionality in the joint state space when tracking multiple targets, 
\cite{khan_mcmc-based_2004} proposes a PF that uses a Markov random field (MRF) to penalize the proximity of particles from different objects.
\cite{hess_discriminatively_2009} describes a multi-target tracking framework based on pseudo-independent log-linear PFs and 
an algorithm to discriminatively train filter parameters. 
\cite{breitenstein_robust_2009} uses a detection-based PF framework with a greedy \mbox{strategy} to associate detection responses.
In contrast to these 
approaches, we focus on using the PF to adaptively estimate the agents' internal goals (e.g., desired location), as well as the position and velocity.

\noindent
{\bf Higher-order models:} Previous works on ``higher-order'' Bayesian models related to tracking include: 
\cite{felsberg2009learning}, which embeds multiple previous states into a large vector for state transition;
\cite{park2012robust}, which is conditioned on multiple previous observations;
\cite{pan2011visual}, which formulates the state transition as a unimodal Gaussian distribution with the mean as a weighted sum of predictions from previous particles. Refer to Sec.~\ref{sec:cmp} for detailed comparisons with our HPF.

\noindent
{\bf Crowd motion models:} Crowd tracking methods use complex motion models
to improve 
tracking large \mbox{crowds}~\cite{ali_floor_2008,rodriguez2009tracking,song_online_2010,kratz2012tracking,yang2012multi}. E.g.,
\cite{ali_floor_2008} uses a floor-fields 
to generate the probability of targets' moving to their nearby regions in extremely crowded scenes, while
\cite{yang2012multi} learns the non-linear motion patterns to assist association of tracklets from detection responses.
Related to our work are trackers using agent-based motion models. E.g.,
\cite{antonini_behavioral_2006} discretizes the velocity space of a pedestrian into multiple areas and models the probability of choosing velocities for tracking, while
\cite{pellegrini_youll_2009} formulates pedestrians' movements
 as an energy optimization problem that factors in navigation and collision avoidance.
\cite{yamaguchi_who_2011} obtains 
insights from the social force approach,
by using both group and destination priors in the pedestrian motion model.
Pedestrian motion is also studied in other fields, e.g. pedestrian dynamics, graphics and robotics~\cite{helbing_social_1995,guy2012least}.
Recently, RVO~\cite{van_den_berg_reciprocal_2011} has also been applied to tracking \cite{jin2012single}.

With these previous approaches~\cite{antonini_behavioral_2006,pellegrini_youll_2009,yamaguchi_who_2011, jin2012single}, agent-based motion model is simply used to predict one step ahead for the state transitions, regardless of its longer-term predictive capabilities.
In contrast, our focus is to fully leverage the capabilities of agent-based motion models, by using longer-term predictions and online estimation of the agent's internal goals without scene priors.

\section{Tracking with particle filters and RVO}
\label{sec:rvotrack}

In this section, we first present the RVO crowd model, and 
integrate it into a multiple-target tracking algorithm using the particle filter.

\subsection{Reciprocal velocity obstacle (RVO)}

The RVO motion model~\cite{van_den_berg_reciprocal_2011} predicts agents' positions and velocities in a 2D ground space, given the information of all agents at the current time step. The predicted agents' positions and velocities are such that they will not lead to collision among the agents. Each agent is first simplified as a disc in 2D space. The formulation of RVO then consists of several parameters, which are used to manipulate each agent's movement, including desired velocity and the radius of agent's disc.

\begin{figure}
        \centering
	\begin{tabular}{c@{}c@{}c@{}c}
	{\footnotesize a)}
          \includegraphics[width=0.35\linewidth]{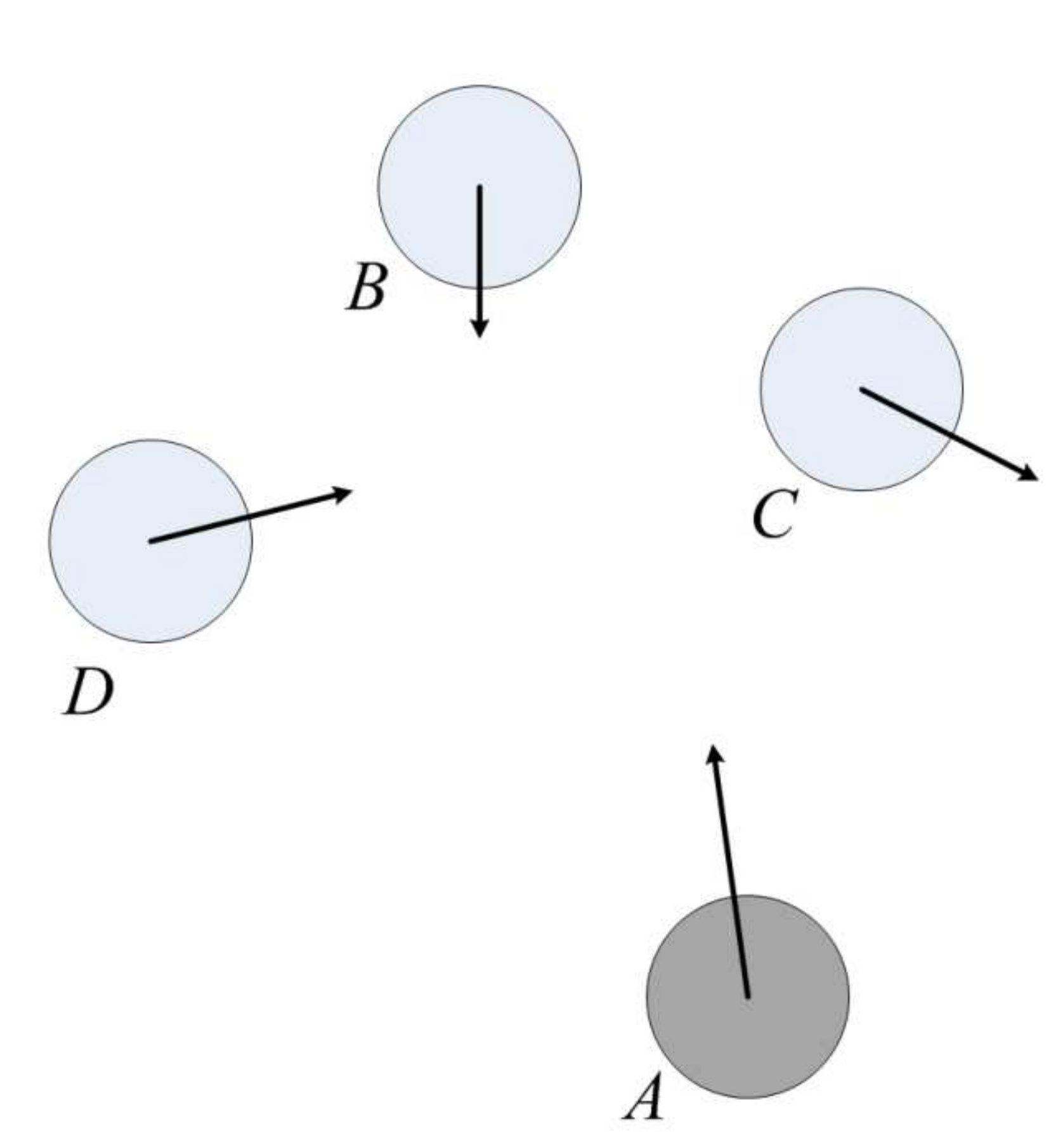} &
          {\footnotesize b)}
          \includegraphics[width=0.35\linewidth]{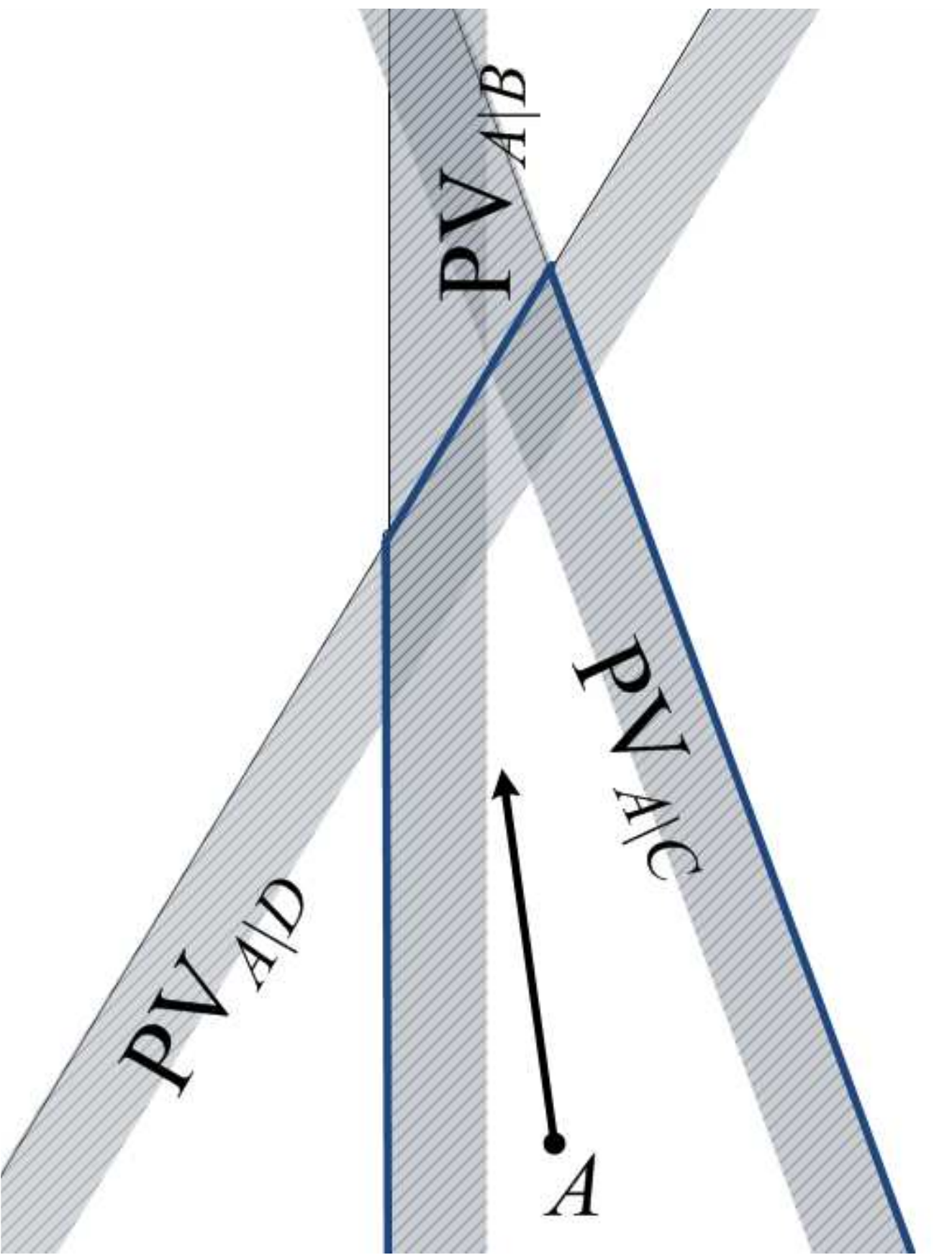}
           \end{tabular}
        \caption{{\small The RVO multi-agent simulation. (a) shows a scenario of four agents in a 2D space, with velocities indicated by arrows. (b) shows the corresponding velocity space. The shaded side of each plane indicates the set of permitted velocities for agent $A$ that avoid collision with each of other agents. The region with bolded lines denotes the set of velocities of agent $A$ that lead to no collisions.}}
        \label{fig:rvo}
\end{figure}

RVO is built on the concept of velocity obstacle ($VO$)~\cite{fiorini_motion_1998}. Let $D(\textbf{p}, r)$ denote an open disc of radius $r$ centered at a 2D position $\textbf{p}$:
\begin{equation}
    D(\textbf{p}, r) = \{ \textbf{q} \mid \parallel \textbf{q} - \textbf{p} \parallel < r \}.
\end{equation}
Given a time horizon $\tau$, we have:
\begin{equation}
    VO^{\tau}_{A \mid B}=\{ \textbf{v} \mid \exists t \in [0,\tau], t\textbf{v} \in D(\textbf{p}_{B}-\textbf{p}_{A}, r_A+r_B)  \}.
\end{equation}

Here $VO^{\tau}_{A \mid B}$ refers to the set of velocities that will lead to the collision of agents $A$ and $B$. Hence, if agent $A$ chooses a velocity that is outside of $VO^{\tau}_{A \mid B}$, $A$ is guaranteed not to collide with $B$ for at least $\tau$ seconds, where $\tau$ is the planning horizon. RVO sets up linear constraints in the velocity space and ensures that $A$'s new velocity will be outside $VO^{\tau}$. Let \textbf{u} be the smallest change required in the relative velocity of $A$ towards $B$ in order to avoid collision. Let $\textbf{v}^{opt}_A$ and $\textbf{v}^{opt}_B$ be the velocities of agents $A$ and $B$, respectively, at the current time step. $\textbf{u}$ can be geometrically interpreted as the vector going from the current relative velocity to the closest point on the $VO$ boundary,
\begin{equation}
    \textbf{u} = (\argmin_{\textbf{v} \in VO^{\tau}_{A \mid B}} \parallel \textbf{v}-(\textbf{v}^{opt}_A-\textbf{v}^{opt}_B) \parallel)-(\textbf{v}^{opt}_A-\textbf{v}^{opt}_B).
\end{equation}

If both agents $A$ and $B$ are implicitly sharing the responsibility to avoid collision, each needs to change their velocity by at least $\frac{1}{2}\textbf{u}$, expecting the other agent to take care of the other half. Therefore, the set of velocities permitted for agent $A$ towards agent $B$ is:
\begin{equation}
    PV^{\tau}_{A \mid B} = \{ \textbf{v}|(\textbf{v}-(\textbf{v}^{opt}_{A}+\frac{1}{2}\textbf{u})) \cdot \textbf{n} \geq 0 \},
\end{equation}
\noindent where $\textbf{n}$ is the normal of the closest point on $VO^{\tau}_{A \mid B}$ to maximize the allowed velocities. In a multi-agent planning situation, this set of permitted velocities for agent $A$ is computed as,
\begin{equation}
PV^{\tau}_A = D(\textbf{0}, v^{max}_A)\cap \bigcap_{B \neq A} PV^{\tau}_{A \mid B},
\end{equation}
\noindent where $v^{max}_A$ is the velocity with the maximum speed that agent $A$ can take. Thus, $A$'s collision-free velocity and position at the next time step can be computed as,
\begin{equation}\label{eq:rvo_evolve}
    \textbf{v}_{t+1} =\argmin_{\textbf{v}\in PV} \parallel \textbf{v}-\textbf{v}_{desire}, \parallel
\end{equation}
\begin{equation}\label{eq:rvo_evolve2}
    \textbf{p}_{t+1}=\textbf{p}_{t}+\textbf{v}_{t+1} \triangle{t},
\end{equation}
\noindent where $\triangle t$ is the time interval between steps and  $\textbf{v}_{desire}$ is the desired velocity. This leads to a convex optimization solution. Figure~\ref{fig:rvo} illustrates an example of RVO. More details can be found in \cite{van_den_berg_reciprocal_2011}.

\subsection{Tracking with particle filters}
\label{sec:RVOandPF}

In our tracking approach, we use an independent particle filter for each person, similar to
\cite{hess_discriminatively_2009}, to track the state of the person.
Each particle filter only estimates the state of a single person, but it can access the previous state estimates of the other pedestrians to infer the velocity using RVO.
Hence the particle filters are loosely coupled throughout the tracking and simulation.

The state representation of a person at time $t$ \mbox{contains} the person's position, velocity, and desired velocity, i.e.,  $\textbf{x}_t = [\textbf{p}_t^T, \textbf{v}_t^T, \textbf{v}_{desire}^T]^T$.
 While $\textbf{p}_t$ and $\textbf{v}_t$ determine the physical properties of the person, $\textbf{v}_{desire}$ represents the intrinsic goal of the person.
In contrast to previous tracking methods using crowd models \cite{pellegrini_youll_2009, yamaguchi_who_2011}, our method automatically performs {\em online estimation} of the intrinsic properties (desired location or desired velocity) during tracking.

In the standard online Bayesian tracking framework, the propagation of state $\textbf{x}_t$ at time $t$ depends only on the previous state (first-order Markov assumption).
The goal of the online Bayesian filter is to estimate the posterior distribution of state $\textbf{x}_t$, given the current observation sequence $\textbf{y}_{1:t}$,
\begin{equation}
    p(\textbf{x}_t | \textbf{y}_{1:t})=\alpha 
    p(\textbf{y}_t | \textbf{x}_t) \int p(\textbf{x}_t | \textbf{x}_{t-1})p(\textbf{x}_{t-1}|\textbf{y}_{1:t-1}) d\textbf{x}_{t-1},
\end{equation}
\noindent where $\textbf{y}_t$ is the observation at time $t$, $\textbf{y}_{1:t}$ is the set of all observations through time $t$, $p(\textbf{x}_t | \textbf{x}_{t-1})$ is the state transition distribution, $p(\textbf{y}_t | \textbf{x}_t)$ is the observation likelihood distribution, and $\alpha$ is the normalization constant.
As the integral does not have a closed-form solution in general, the particle filter \cite{doucet2001introduction}
approximates the posterior using a set of weighted samples $\mathcal{X}_t = \{ (w^{[m]}_t, \textbf{x}^{[m]}_t) \}_{m=1:M}$, where each $\textbf{x}^{[m]}_t$ is an instantiation of the process state, known as a particle, and $w^{[m]}_t$ is the corresponding weight. Under this representation, the Monte Carlo approximation can be given as
\begin{equation}
    p(\textbf{x}_t | \textbf{y}_{1:t})\thickapprox \alpha \hspace{3 pt} p(\textbf{y}_t | \textbf{x}_t) \sum^M_{m=1} w^{[m]}_{t-1} p(\textbf{x}_t | \textbf{x}^{[m]}_{t-1}),
\end{equation}
\noindent where $M$ is the number of particles.

For the transition density, 
we model the propagation of the velocity and position as the RVO prediction with additive Gaussian noise, and use a simple diffusion process for the desired velocity.
Thus, the transition density is
\begin{equation}\label{eq:transit}
 p(\textbf{x}_{t+1} | \textbf{x}_{t})=\mathcal{N}(\begin{bmatrix} \textbf{p}_{t}+\mathcal{R}(\textbf{X}_t) \Delta t \\ \mathcal{R}(\textbf{X}_t) \\ \textbf{v}_{desire} \end{bmatrix}, \Gamma),
\end{equation}
where we represent the process of computing the predicted velocity using RVO in Eq.~\ref{eq:rvo_evolve} as $\mathcal{R}(\textbf{X}_t)$.
$\textbf{X}_t$ is the set of posterior means for all agents at time $t$,
and $\Gamma$ is a diagonal covariance matrix of the noise.

\section{Tracking with higher-order particle filters}
\label{sec:HPF}

Due to the modeling of person-to-person reciprocal interactions, RVO can reliably predict the crowd configuration over longer times.
However, the particle filter from Section \ref{sec:RVOandPF} uses a first-order Markov assumption, and hence the crowd model only predicts ahead one time step. Aiming to take better advantage of the longer-term predictive capabilities of RVO, in this section, we derive a higher-order particle filter (HPF) that uses multiple predictions, starting from past states.

\subsection{Higher-order particle filter}
In the HPF, we assume that the state process evolves according to a $K$th-order Markov model, with transition distribution $p(\textbf{x}_{t}|\textbf{x}_{t-1:t-K})$,
which is a mixture of individual transitions from each previous state $\textbf{x}_{t-j}$,
\begin{equation}\label{eq:assump}
p(\textbf{x}_{t}|\textbf{x}_{t-1:t-K}) = \sum_{j=1}^{K} \pi_{j} p_{j}(\textbf{x}_{t}|\textbf{x}_{t-j}),
\end{equation}
\noindent where $\pi_j$ are  mixture weights, $\sum_{j}\pi_{j}=1$, and $p_j(\textbf{x}_t | \textbf{x}_{t-j})$ is the $j$-step ahead transition distribution, i.e., the prediction from previous state $\textbf{x}_{t-j}$ to time $t$.
For the $K$th-order model, the predictive distribution is
\begin{align}
\nonumber
\lefteqn{p(\textbf{x}_{t} | \textbf{y}_{1:t-1}) = \int p(\textbf{x}_{t}|\textbf{x}_{t-1:t-K})p(\textbf{x}_{t-1:t-K} | \textbf{y}_{1:t-1}) d\textbf{x}_{t-1:t-K}}
\\
\nonumber
&=
\int \left[\sum_{j=1}^{K} \pi_j p_j(\textbf{x}_t|\textbf{x}_{t-j})\right]p(\textbf{x}_{t-1:t-K} | \textbf{y}_{1:t-1})d\textbf{x}_{t-1:t-K}
\\
&=
\sum_{j=1}^K \pi_{j} \int p_{j}(\textbf{x}_{t}|\textbf{x}_{t-j}) p(\textbf{x}_{t-j}|\textbf{y}_{1:t-1}) d\textbf{x}_{t-j},
\label{eqn:deriv1}
\end{align}
%
where Eq.~\ref{eqn:deriv1} follows from swapping the summation and integral, and marginalizing out the state variables $\textbf{x}_{t-k}$, for $k\neq j$.
Note that the posterior of the $j$th previous state $p(\textbf{x}_{t-j}|\textbf{y}_{1:t-1})$ depends on some ``future'' observations $\{\textbf{y}_{t-j+1}, \cdots, \textbf{y}_{t-1}\}$.
As we want to leverage the longer-term predictive capabilities of our crowd model, we assume that these ``future'' observations are unseen, and hence $p(\textbf{x}_{t-j}|\textbf{y}_{1:t-1})\approx p(\textbf{x}_{t-j}|\textbf{y}_{1:t-j})$. The motivation for the approximation is two-fold: 1) it makes HPF completely online, by avoiding backtracking needed for smoothing PFs; 2) it reduces the effects of bad observations (e.g. if occlusion occurs at t-1, predictions from t-2 can jump over it). 
Substituting it into Eq.~\ref{eqn:deriv1} yields an approximate predictive distribution as
\begin{equation}
p(\textbf{x}_{t} | \textbf{y}_{1:t-1}) \approx \sum_{j=1}^{K}  \pi_{j} p_j(\textbf{x}_{t}|\textbf{y}_{1:t-j}),
\end{equation}
which is a weighted sum of individual $j$-step ahead predictive distributions from the state posterior at time $t-j$,
\begin{align}
p_j(\textbf{x}_{t}|\textbf{y}_{1:t-j}) &= \int p_j(\textbf{x}_{t}|\textbf{x}_{t-j})p(\textbf{x}_{t-j}|\textbf{y}_{1:t-j})d\textbf{x}_{t-j}.
\end{align}
The $j$-step ahead predictive distribution is approximated with particles from time $t-j$,
\begin{align}
p_j(\textbf{x}_{t}|\textbf{y}_{1:t-j}) &\approx \sum_{m} p_j(\textbf{x}_{t}|\textbf{x}^{[m]}_{t-j}) p(\textbf{x}^{[m]}_{t-j}|\textbf{y}_{1:t-j}).
\end{align}
Associated with the $j$-step ahead predictor is a corresponding posterior distribution, which is further conditioned on the current observation $\textbf{y}_t$,
	\begin{align}
	\lefteqn{
	p_j(\textbf{x}_t|\textbf{y}_{1:t-j},\textbf{y}_t)
	= \frac{p(\textbf{y}_t|\textbf{x}_t) p_j(\textbf{x}_t|\textbf{y}_{1:t-j})}
	{p_j(\textbf{y}_t|\textbf{y}_{1:t-j})}
	}
	\\
	&\approx \tfrac{1}{\alpha_j} 	p(\textbf{y}_{t} | \textbf{x}_{t})
\sum_m p_j(\textbf{x}_{t}|\textbf{x}^{[m]}_{t-j}) p(\textbf{x}^{[m]}_{t-j}|\textbf{y}_{1:t-j}),
	\end{align}
where $\alpha_j=p_j(\textbf{y}_t|\textbf{y}_{1:t-j})$ is the likelihood of observing $\textbf{y}_t$ using the $j$-step ahead predictor,
	\begin{align}
	p_j(\textbf{y}_t|\textbf{y}_{1:t-j}) &= \int p(\textbf{y}_t|\textbf{x}_t) p_j(\textbf{x}_t|\textbf{y}_{1:t-j}) d\textbf{x}_t
	\\
	&\approx \sum_m p(\textbf{y}_{t} | \textbf{x}^{[m]}_{t}) p_j(\textbf{x}^{[m]}_{t} | \textbf{y}_{1:t-j}).
	\end{align}
	
The posterior distribution can now be approximated as
	\begin{align}
	p(\bx_{t}|\by_{1:t})
	&=
	\frac{p(\by_t|\bx_t)p(\bx_t|\by_{1:t-1})}{\int p(\by_t|\bx_t)p(\bx_t|\by_{1:t-1})d\bx_t}
	\\
	&\approx
	\frac{p(\by_t|\bx_t)\sum_j \pi_j p_j(\bx_t|\by_{1:t-j})}{\int p(\by_t|\bx_t)\sum_j \pi_j p_j(\bx_t|\by_{1:t-j})d\bx_t}
	\\
	&=
	\frac{\sum_j \pi_j p(\by_t|\bx_t) p_j(\bx_t|\by_{1:t-j})}{\sum_j \pi_j\int p(\by_t|\bx_t) p_j(\bx_t|\by_{1:t-j})d\bx_t}
	\\
	&=
	\frac{\sum_j \pi_j \frac{p_j(\by_t|\by_{1:t-j})}{p_j(\by_t|\by_{1:t-j})}p(\by_t|\bx_t) p_j(\bx_t|\by_{1:t-j})}{\sum_j \pi_j p_j(\by_t|\by_{1:t-j})}
	\\
	&=
	\frac{\sum_j \pi_j p_j(\by_t|\by_{1:t-j})p_j(\bx_t|\by_{1:t-j},\by_t)}{\sum_j \pi_j p_j(\by_t|\by_{1:t-j})}
	\end{align}
Hence,
the approximate posterior distribution is
a weighted sum of the $j$-step ahead posteriors,
\begin{align}\label{eq:hpf_final}
p(\textbf{x}_{t}|\textbf{y}_{1:t}) \approx \sum_j^K \lambda_j p_j(\textbf{x}_{t}|\textbf{y}_{1:t-j},\textbf{y}_{t}),
\end{align}
where the weight for each individual posterior is
\begin{equation}
\label{eqn:lambda}
\lambda_j = \frac{\pi_{j} p_j(\textbf{y}_{t} | \textbf{y}_{1:t-j}) }{ \sum_j \pi_{j} p_j(\textbf{y}_{t} | \textbf{y}_{1:t-j}) }.
\end{equation}

Finally, substituting in the particle filter \mbox{approximation,}
\begin{align}
p(\bx_t|\by_{1:t})&\propto
\sum_j^K \lambda_j p(\textbf{y}_{t} | \textbf{x}_{t})
\sum_m p_j(\textbf{x}_{t}|\textbf{x}^{[m]}_{t-j}) p(\textbf{x}^{[m]}_{t-j}|\textbf{y}_{1:t-j}),
\end{align}

\begin{align}
p_j(\textbf{y}_{t} | \textbf{y}_{1:t-j}) &= \int p(\textbf{y}_{t} | \textbf{x}_{t}) p_j(\textbf{x}_{t} | \textbf{y}_{1:t-j})  d\textbf{x}_{t} \\
&\approx \sum_m p(\textbf{y}_{t} | \textbf{x}^{[m]}_{t}) p_j(\textbf{x}^{[m]}_{t} | \textbf{y}_{1:t-j}).
\end{align}

\begin{figure}
  \centering
  \includegraphics[width=4in]{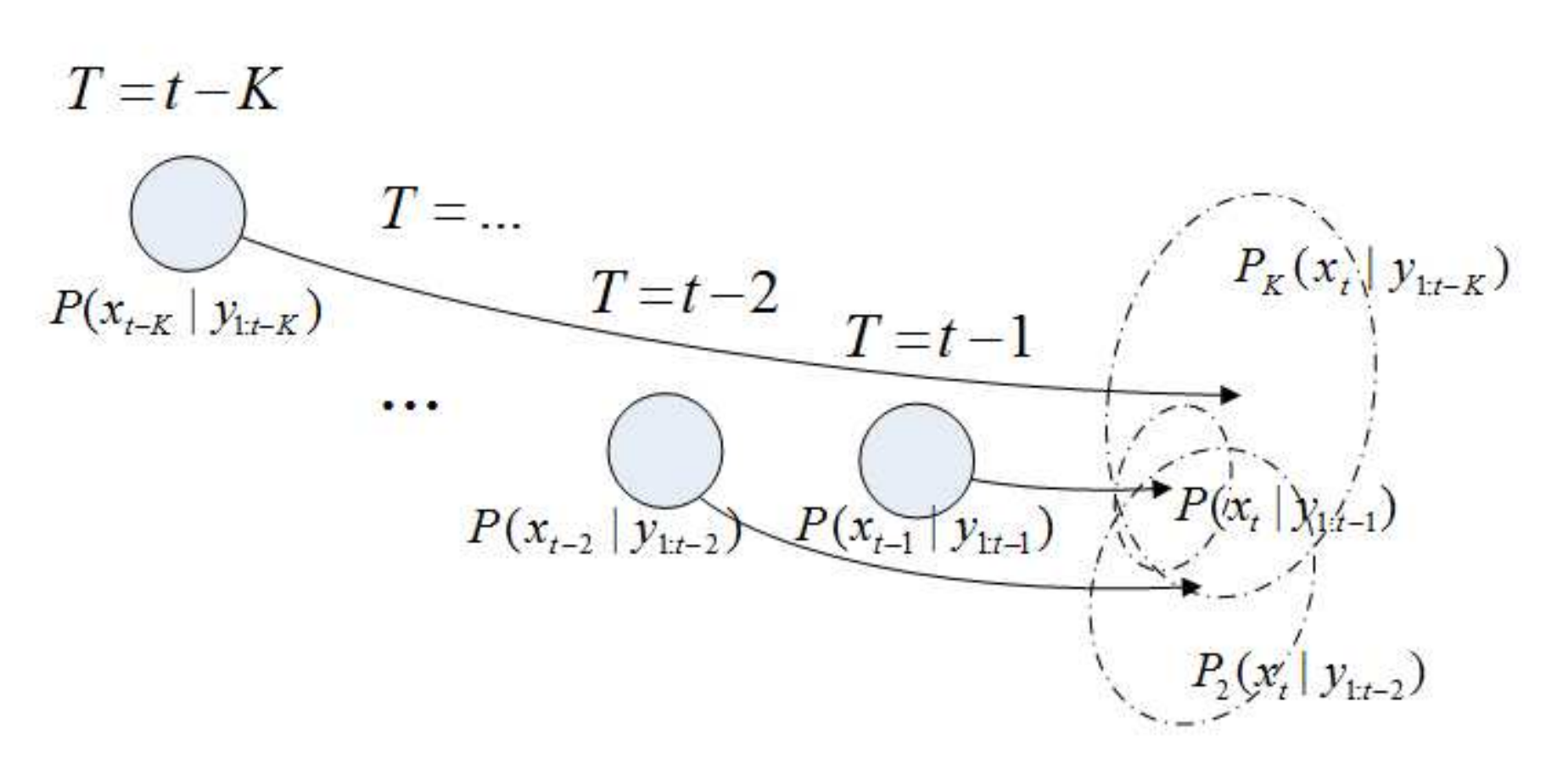}
  \caption{{\small The basic principle of the higher-order particle filter (HPF). Each circle represents the posterior of the previous state at time steps $t-1$, $t-2$, ..., $t-K$. Each dashed ellipse is the predictive distribution evolved from the posterior of the previous state. The resulting posterior of the current time step is the weighted sum of these predictive distributions (Eq.~\ref{eq:hpf_final}).}}
\label{fig:hpf}
\end{figure}

In summary, the posterior of HPF is a weighted sum of standard particle filters, where each filter computes a $j$-step ahead prediction from time $t-j$ to $t$.
The weights are proportional to the likelihood of the current observation for each filter, and hence, some modes of the posterior will be discounted if they are not well explained by the current observation.
The overview of HPF is illustrated in Figure~\ref{fig:hpf} and the pseudo-code is in Algorithm~\ref{algo}.

Note that HPF is computationally efficient, as we may reuse the previous particles (i.e., $\textbf{x}^{[j,m]}_t$ in Algorithm~\ref{algo}) and the $j$-step ahead predictions can be computed recursively from the previous predictions.

\begin{algorithm}
\caption{{\small The Higher-Order Particle Filter.}}
\begin{small}
\begin{algorithmic}[1]
\STATE $\mathcal{X}_t = \overline{\mathcal{X}}_t = \emptyset$.
\FORALL{time step $t \in \{1...T\}$}
    \FORALL  {$j \in \{1, ...,K \}$}
    \STATE \COMMENT{* calculate individual posterior from time $t-j$ *}
        \FORALL  {$m \in \{1...M\}$}
            \STATE $\textbf{v}_{t} = \mathcal{R}(\textbf{X}_{t-j})$ .  \hfill\COMMENT {* RVO prediction *}
            \STATE $\overline{\textbf{x}}_{t} = [(\textbf{p}_{t-j} +  \textbf{v}_{t} \triangle t)^T, \textbf{v}_{t}^T, \textbf{v}_{desire}^T]^T$.
            \STATE $p(\textbf{x}_t | \textbf{x}_{t-j})\sim \mathcal{N}(\overline{\textbf{x}}_{t}, \Gamma)$.
            \STATE $\textbf{x}^{[j,m]}_t \sim p(\textbf{x}_t | \textbf{x}^{[m]}_{t-j}) $. \hfill\COMMENT{* update particle *}
            \STATE $w^{[j,m]}_{t}=w^{[m]}_{t-j} p(\textbf{y}_t | \textbf{x}^{[j,m]}_{t})$. \hfill\COMMENT{* update weight *}
        \ENDFOR
    \ENDFOR

  \STATE \COMMENT{* calculate posterior *}
    \FORALL  {$j \in \{1, ...,K \}$}
        \STATE $\lambda_j = \mathrm{Normalize}(\pi_j \sum_{m=1}^M w^{[j,m]}_{t})$.
        \STATE $w^{[j,1:M]}_{t} = \mathrm{Normalize}(w^{[j,1:M]}_{t})$.
        \FORALL {$m \in \{1...M\}$}
            \STATE $w^{[j,m]}_{t}=\lambda_j  w^{[j,m]}_{t}$.
            \STATE $\overline{\mathcal{X}}_t = \overline{\mathcal{X}}_t \cup \{ (w^{[j,m]}_{t}, \textbf{x}^{[j,m]}_t) \}$.
        \ENDFOR
    \ENDFOR
    \STATE \COMMENT{* resample and select most weighted $M$ particles *}
	\STATE $\mathcal{X}_t = \mathrm{Resample}(\overline{\mathcal{X}}_t,M)$.
\ENDFOR
\STATE NOTE: $\mathcal{R}(\textbf{X}_{t-j})$ predicts the velocity at time $t$ from the state $\textbf{x}_{t-j}$
by recursively applying a single step of RVO $j$ times.
\end{algorithmic}
\end{small}
\label{algo}
\end{algorithm}

\subsection{Comparison with previous higher-order models}
\label{sec:cmp}

Different from previous methods, our HPF formulation assumes the state transition a mixture of individual transitions from multiple prior time steps. Felsberg et al.~\cite{felsberg2009learning} embed multiple previous states into a vector, resulting in a transition density that is a product of individual transitions terms, while ours is a mixture distribution. Park et al.~\cite{park2012robust} use an appearance model conditioned on multiple previous observations, while we use a first-order appearance model. Both \cite{felsberg2009learning,park2012robust} are not based on PFs. Our work is closely related to \cite{pan2011visual}, which formulates a higher-order PF where the transition model is a unimodal Gaussian distribution with the mean as a weighted sum of predictions from previous particles. In \cite{pan2011visual}, the influence of previous states depends on their trained transition weights rather than the observation likelihood, and each particle depends only on its parent particles. In contrast, our higher-order PF models the state transition as a multimodal mixture distribution and the importance weights rely on both observations and transition weights (Eq.~\ref{eqn:lambda}). Specifically, previous states with high likelihood in predicting the current observation impose more effects on the posterior, which is useful for tracking through occlusion. Moreover, all particles of the previous states are used to form the posterior, which increases diversity of particles (Algorithm~\ref{algo}).

\section{Experiments}
\label{sec:results}

We 
evaluate our tracking algorithm in two experiments: 
1) prediction of pedestrians' motion from ground-truth trajectories; 
2) multi-pedestrian tracking in video with low fps.

\subsection{Dataset and models}
\label{sec:models}
We use the same crowd datasets used in ~\cite{pellegrini_youll_2009,yamaguchi_who_2011} to study real-world interactions among pedestrians and evaluate motion models. We use the \emph{Hotel} sequences from~\cite{pellegrini_youll_2009} and the \emph{Zara01}, \emph{Zara02} and \emph{Student} sequences from~\cite{lerner2007crowds}. All sequences are captured at 2.5 fps and annotated at 0.4s intervals. \emph{Hotel} is captured from an aerial view, while \emph{Zara01} and \emph{Zara02} are both side views,
and \emph{Student} is captured 
from an oblique view. All of them include adequate interactions of pedestrians (e.g. collision avoidance). 

In our framework, we transform pedestrians’ feet positions in the image to coordinates in the ground space, as is the case in many recent tracking papers~\cite{andriyenko_discrete-continuous_2012,pellegrini_youll_2009,yamaguchi_who_2011}; pedestrians’ motion can be estimated and described using pedestrian motion model such as RVO and LTA for the sake of accuracy. Tracking approaches using ground-space information become popular due to advances in estimating the transformation between the world space and the image space, e.g. Fig.7 in ~\cite{pellegrini_youll_2009} uses a moving camera at eye-level. Here the geometric transformation is enough for static camera. Our datasets follow~\cite{pellegrini_youll_2009,yamaguchi_who_2011}, which are all general surveillance videos. Except Hotel, these datasets are captured with oblique views.

In addition to RVO, we also consider three other agent-based motion models, LTA~\cite{pellegrini_youll_2009}, ATTR and ATTRG~\cite{yamaguchi_who_2011}.
LTA and ATTR predict the pedestrians' velocities by optimizing energy functions,
which contain terms for collision avoidance, attraction 
towards the goals and desired speed. ATTRG adds a pedestrian grouping prior to ATTR. The source codes are acquired from the authors of ~\cite{pellegrini_youll_2009,yamaguchi_who_2011}.
We also consider the baseline  constant velocity model (LIN).

In order to make a direct comparison between motion models
in the same framework, we combine each of these motion models with the identical particle filter (PF), described in Sec.~\ref{sec:RVOandPF}, by replacing  RVO with others. 
The state representation of the PF includes the agent's physical properties (position and velocity) and internal goals (desired velocity), as in Sec.~\ref{sec:RVOandPF}.
The desired velocity is initially set to the initial velocity and
evolves independently as the tracker runs.  These online adaptive models are denoted as LTA+, ATTR+, ATTRG+, and RVO+.
We also test a version of the PF where the desired velocity does not evolve along with position and velocity, and is fixed as the initial velocity (denoted as LIN, LTA, ATTR, ATTRG, and RVO).

In addition to PF, we also test RVO+ with our proposed HPF (Sec.~\ref{sec:HPF}).
We compare against the higher-order particle filter from \cite{pan2011visual} (denoted as pHPF), which models the transition density as Gaussian where the mean is the weighted sum of  predictions from previous particles.  We use both RVO and LIN with pHPF.
The higher-order models of \cite{felsberg2009learning,park2012robust} are not based on PFs, so we do not consider them in experiments.

\noindent \textbf{Parameter selection:} As in \cite{pellegrini_youll_2009, yamaguchi_who_2011}, the parameters of \mbox{LTA}, ATTR and RVO are trained using the genetic algorithm on an independent dataset \emph{ETH}.
The hyperparameters of HPF ($K, \pi_j$) and pHPF (weights from each past particle) are also trained on \emph{ETH} from~\cite{pellegrini_youll_2009}.
For HPF, the resulting settings are $K=2$\footnote{Since we test on low fps (2.5fps) data, compared with normal fps (25fps), the locations and velocities of pedestrians may change significantly between 3 frames.  From our experiments, $K=2$ turns out to be a reasonable longer-term duration to predict ahead.},
$\pi_j=\{0.91, 0.09\}$\footnote{The ratio between $\pi_1$ and $\pi_2$ suggests that the prediction from 2-steps ahead should have more than 10x higher likelihood in order to override the prediction from 1-step ahead (Eq.~\ref{eqn:lambda})}.
Outside of this training, no scene prior or destination information is given, since we assume that the tracker can estimate these online. 

\begin{table}[ht]
\hspace{-0.5in}
\centering
\caption{{\small Mean prediction errors of different motion models.  
The bold numbers are the best in each column.}}

\begin{footnotesize}
\begin{tabular}{|c|ccc|ccc|ccc|c|}
\hline
 & \multicolumn{3} {|c|}{\emph{Zara01}} & \multicolumn{3} {|c|}{\emph{Zara02}} & \multicolumn{3} {|c|}{\emph{Student}} &
 \\ \cline{2-10}
   & L=5 & L=15 & L=30 & L=5 & L=15 & L=30 & L=5 & L=15 & L=30 & avg.\\
   \hline
LIN & 0.35 & 0.69 & 0.78 & 0.38 & 0.79 & 0.89 & 0.39 & 0.84 & 1.00 & 0.68 \\
\hline
LTA & 0.34 & 0.64 & 0.72 & 0.36 & 0.70 & 0.80 & 0.42 & 0.84 & 0.98 & 0.64 \\
LTA+ & 0.37 & 0.73 & 0.82 & 0.37 & 0.77 & 0.89 & 0.42 & 0.91 & 1.08 & 0.71 \\
\hline
ATTR & 0.43 & 0.85 & 0.94 & 0.41 & 0.80 & 0.92 & 0.55 & 1.06 & 1.22 &  0.80 \\
ATTR+ & 0.35 & 0.69 & 0.77 & 0.36 & 0.72 & 0.82 & 0.42 & 0.84 & 0.98 & 0.66 \\
ATTRG & 0.42 & 0.80 & 0.88 & 0.41 & 0.81 & 0.93 & 0.55 & 1.06 & 1.22 & 0.79 \\
ATTRG+ & 0.36 & 0.67 & 0.74 & 0.36 & 0.70 & 0.80 & 0.46 & 0.88 & 1.02 & 0.67 \\
\hline
RVO & 0.34 & 0.66 & 0.74 & 0.34 & 0.72 & 0.84 & 0.39 & 0.83 & 0.98 & 0.65 \\
RVO+ & 0.31 & 0.61 & 0.69 & 0.34 & 0.69 & 0.80 & 0.39 & 0.82 & 0.97 & 0.62 \\
\hline
pHPF (LIN) & 0.41 & 0.79 & 0.88 & 0.43 & 0.86 & 1.00 & 0.46 & 0.95 & 1.11 & 0.77 \\
  pHPF(RVO+) &	0.37	& 0.68 &	0.76 &	0.38	& 0.76	&0.88 &	0.44	&0.90	& 1.05	& 0.69\\
HPF (RVO+) & \textbf{0.25} & \textbf{0.52} & \textbf{0.60} & \textbf{0.26} & \textbf{0.61} & \textbf{0.69} & \textbf{0.31} & \textbf{0.72} & \textbf{0.86} & \textbf{0.54} \\
\hline
\end{tabular}
\end{footnotesize}
\label{tab:prediction}
\end{table}

\subsection{Online estimation and motion prediction}

In the first experiment, we evaluate the motion \mbox{models} mentioned in Sec.~\ref{sec:models} and online estimation of desired velocities on a ground-truth prediction problem.
For each dataset, the following two-phase experiment starts at every 16th time-step of the sequence.
(\emph{Learning})
the PF iterates 
using the ground-truth positions as observations 
for 10 consecutive time steps;
(\emph{Prediction}) after 10  time steps,
the underlying motion model predicts
(without further observations)
the person's trajectory for at most 30 time steps.
We evaluate the accuracy of the motion models' predictions using mean errors, i.e., the distance from ground-truth, averaged over all trajectories in each dataset.

\begin{figure}
        \centering
        \begin{subfigure}[b]{0.4\textwidth}
                \centering
                \includegraphics[width=\textwidth]{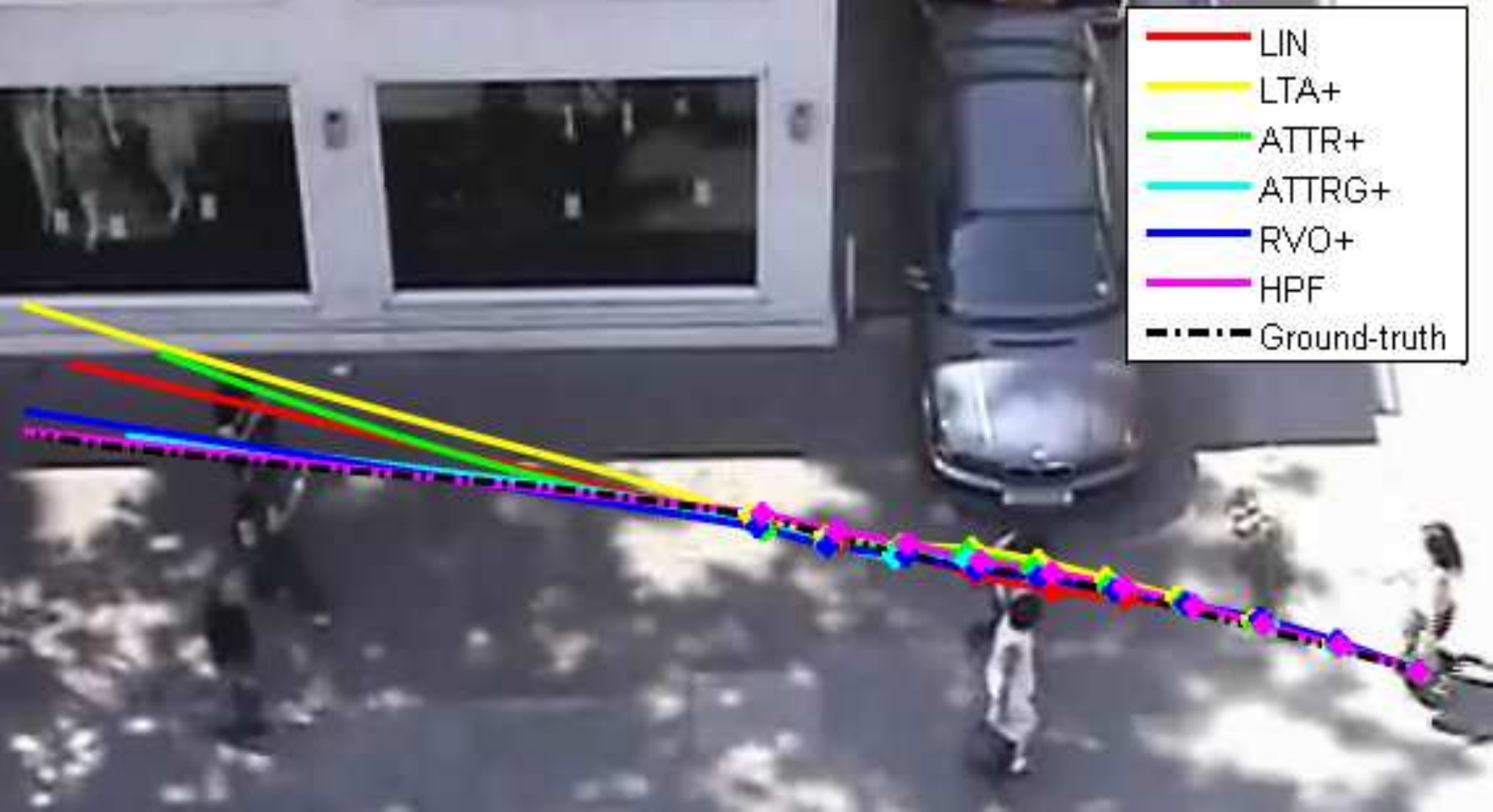}
        \end{subfigure}
        \hspace*{0.01 in}
        \begin{subfigure}[b]{0.4\textwidth}
                \centering
                \includegraphics[width=\textwidth]{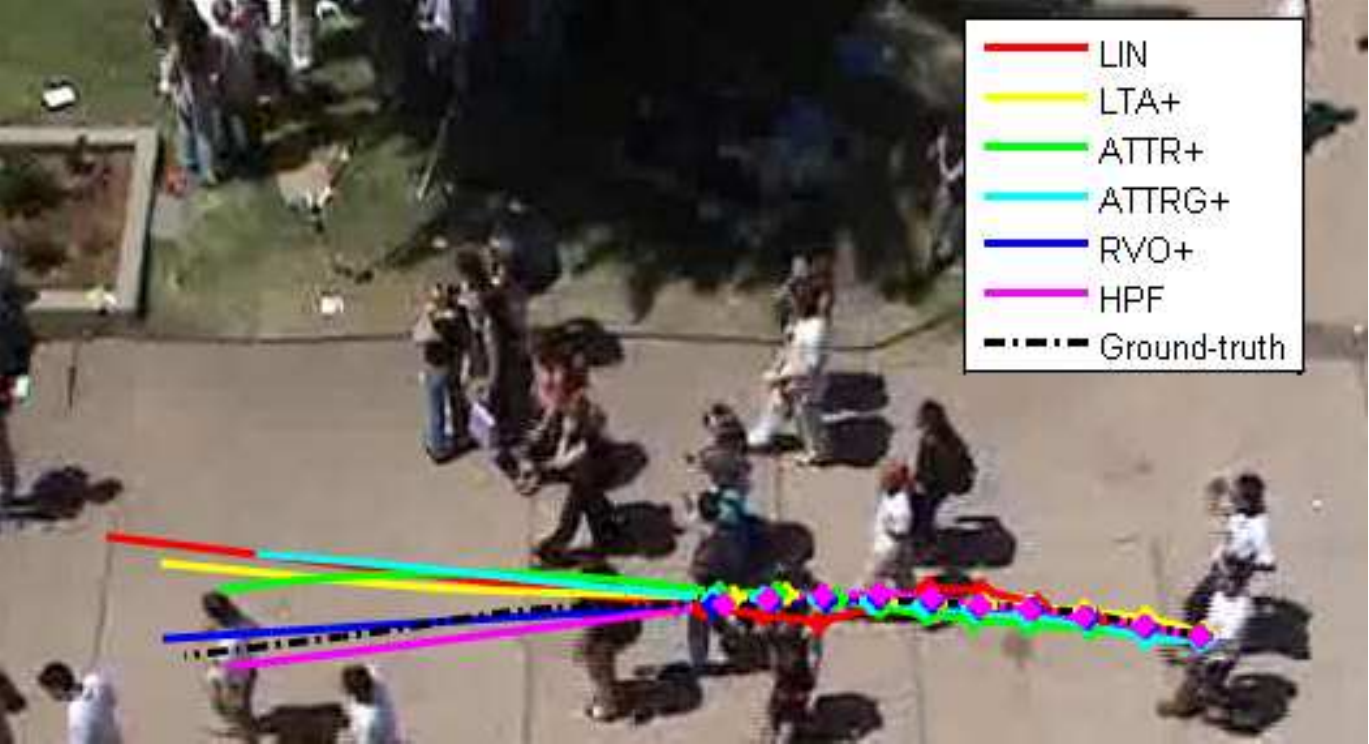}
        \end{subfigure}
        \caption{{\small Prediction on \emph{Zara01} and \emph{Student}.
        The markers indicate the points used in the learning phase, while the solid line is the subsequent prediction.
        %
        The predicted trajectories of RVO+ and HPF 
        are adapted through the learning phase,
        and 
        are closest to the ground-truth.
        }}
        \label{fig:pred_zara}
\end{figure}

Table~\ref{tab:prediction} reports the errors of predicting $L=\{5,15,30\}$ time steps on each dataset, as well as the average error over all datasets.
Looking at the prediction performance without online estimation of desired velocities, LTA and RVO have similar error (0.64/0.65).  However, due to the absence of the goals, ATTR and ATTRG do not perform as well. With the desired velocities adapted, the prediction errors decrease for most motion models (ATTR+, ATTRG+, and RVO+). This suggests that these crowd motion models rely on online estimating goals to produce accurate predictions. Since the behavior prediction of LTA relies on the velocity at the previous time step, the desired velocity estimation (LTA+) does not improve prediction. Among motion models, RVO+ has the lowest prediction error (0.62).
Finally, using the HPF with RVO+ improves the prediction over the standard PF (RVO+) for all prediction lengths, with the average error reduced to 0.54 (about $\sim$15\% lower).
Figure~\ref{fig:pred_zara} shows two examples comparing RVO+ with related models.

\subsection{Multiple-person tracking}

We have evaluated our framework for tracking pedestrians
in videos with low frame rate. 
%

\textbf{Setup.}~The observation likelihood is modeled as
$p(\textbf{y}_t|\textbf{x}_t) \propto \mathrm{HSV}(\textbf{p}_t, \textbf{y}_t) \mathrm{Detect}(\textbf{p}_t, \textbf{y}_t)$.
The first term, $\mathrm{HSV}(\textbf{p}_t, \textbf{y}_t)$, measures the color histogram similarity between the observed template and the appearance template centered at location $\textbf{p}_t$ in frame $\textbf{y}_t$, i.e., $\mathrm{HSV}(\textbf{p}_t, \textbf{y}_t) \propto \exp(- B(\textbf{p}_t, \textbf{y}_t)^2/2\sigma^2)$, where $B(\textbf{p}_t, \textbf{y}_t)$ measures the \mbox{Bhattacharyya} distance between the observed template and the appearance template and $\sigma$ is the variance paramter.
The second term, $\mathrm{Detect}(\textbf{p}_t, \textbf{y}_t)$, is the detection score at  $\textbf{p}_t$ of $\textbf{y}_t$ from a HOG-based head-shoulders detector~\cite{dalal2005histograms}\footnote{We use a similar training/test sets as \cite{yamaguchi_who_2011} for cross validation.
Each dataset is divided into two halves with the detector trained on one half and tested on the other.}.

We use the same experimental protocol described in \cite{yamaguchi_who_2011} for 2.5 fps video. The tracker starts at every 16th frame and runs for at most 24 subsequent frames, as long as the ground truth data exists for those frames in the scene.  The tracker is initialized with the ground-truth positions, and receives no additional future or ground-truth information after initialization.
No scene-specific prior or destination information are used since these are estimated online using our framework.

We evaluate tracking results using successful tracks (ST) and ID switches (IDS), as defined in~\cite{yamaguchi_who_2011}.
A successful track is recorded if the tracker stays within 0.5 meter from its own ground-truth after $N=\{16,24\}$ steps\footnote{We use $N = \{8, 16\}$ for \emph{Hotel} since it contains shorter trajectories.}.
A track that is more than 0.5 meter away from its ground-truth position is regarded as lost, while a track that is within 0.5 meter from its own ground-truth but closer to another person in the ground-truth is considered an ID switch (IDS).  The number of ST and IDS are recorded for each dataset.  We also report the total numbers using the ``short'' interval ($N=8$ for \emph{Hotel} and $N=16$ for \emph{Zara01}, \emph{Zara02}, and \emph{Student}) and ``long'' ($N=16$ for \emph{Hotel} and $N=24$ for others).

\begin{table*}[ht]\centering
\caption{{\small Comparison on the number of successful tracks (larger the better) and ID switches (smaller the better).}}
\begin{footnotesize}

\hspace*{-7 em}
\begin{tabular}{|@{\hspace{0.05in}}c@{\hspace{0.05in}}|c@{\hspace{0.05in}} c@{\hspace{0.05in}} c@{\hspace{0.05in}} c@{\hspace{0.05in}}|c@{\hspace{0.05in}} c@{\hspace{0.05in}} c@{\hspace{0.05in}} c@{\hspace{0.05in}}|c@{\hspace{0.05in}} c@{\hspace{0.05in}} c@{\hspace{0.05in}} c@{\hspace{0.05in}}|c@{\hspace{0.05in}} c@{\hspace{0.05in}} c@{\hspace{0.05in}} c@{\hspace{0.05in}}|c@{\hspace{0.05in}} c@{\hspace{0.05in}} c@{\hspace{0.05in}} c@{\hspace{0.05in}}|}

\hline
 & \multicolumn{4} {|c|}{\emph{Zara01}} & \multicolumn{4} {|c|}{\emph{Zara02}} & \multicolumn{4} {|c|}{\emph{Student}}	 & \multicolumn{4} {|c|}{\emph{Hotel}} & \multicolumn{4} {|c|}{Total}
 \\
 \cline{2-21}
 & \multicolumn{2}{|c}{ST} & \multicolumn{2}{c|}{IDS} & \multicolumn{2}{|c}{ST} & \multicolumn{2}{c|}{IDS} & \multicolumn{2}{|c}{ST} & \multicolumn{2}{c|}{IDS} & \multicolumn{2}{|c}{ST} & \multicolumn{2}{c|}{IDS} & \multicolumn{2}{|c}{ST} & \multicolumn{2}{c|}{IDS}
 \\
 \cline{2-21}
  	  & {\footnotesize N=16} & {\footnotesize N=24} & {\footnotesize N=16} & {\footnotesize N=24}    & {\footnotesize N=16} & {\footnotesize N=24} & {\footnotesize N=16} & {\footnotesize N=24}     & {\footnotesize N=16} & {\footnotesize N=24} & {\footnotesize N=16} & {\footnotesize N=24}    & {\footnotesize N=8} & {\footnotesize N=16} & {\footnotesize N=8} & {\footnotesize N=16} & {\footnotesize short} & {\footnotesize long} & {\footnotesize short} & {\footnotesize long} \\
  \hline  	
  LIN & 149 & 57 & 7 & 2 & 343 & 178 & 24 & 13 & 414 & 197 & 62 & 21 & 457 & 149 & 13 & 6 & 1363 &	581 &	106 &	42  \\
  \hline
  LTA & 146 & 54 & 9 & 1 & 315 & 159 & 16 & 4 & 338 & 162 & 50 & 22 & 457 & 153 & 10 & 3 & 1256	& 528	& 85	& 30  \\
  LTA+ & 154 & 60 & 6 & 0 & 331 & 185 & 20 & 9 & 391 & 185 & 59 & 22 & 465 & 148 & 14 & 2 & 1341 &	578 &	99 &	33  \\
  \hline
  ATTR & 129 & 38 & 9 & 1 & 307 & 170 & 16 & 3 & 276 & 107 & 34 & 12 & 462 & 144 & 12 & 4 & 1174	& 459 &	71 &	20  \\
  ATTR+ & 149 & 67 & 7 & 2 & 339 & 185 & 23 & 7 & 394 & 199 & 49 & 24 & 457 & 145 & 8 & 3 & 1339	& 596 &	87	& 36  \\
 ATTRG & 130 & 39 & 9 & 1 & 300 & 162 & 14 & 3 & 284 & 109 & 36 & 14 & 463 & 144 & 9 & 4 & 1177	& 454 &	68 &	22 \\
  ATTRG+ & 155 & 60 & 9 & 3 & 316 & 180 & 14 & 13 & 381 & 174 & 50 & 24 & 467 & 145 & 12 & 2 & 1319	& 559	& 85	& 42 \\
  \hline
  RVO & 171 & 75 & 4 & 2 & 365 & 194 & 22 & 7 & 412 & 202 & 50 & 25 & 451 & 147 & 7 & 3 & 1399	& 618 &	83 &	37  \\
  RVO+ & 173 & 76 & 5 & 2 & 383 & 209 & 28 & 15 & 419 & 208 & 60 & 25 & 474 & 162 & 5 & 4 & 1449	& 655	& 98 &	46   \\
  \hline
  pHPF(LIN)& 124 & 40 & 9 & 0 & 293 & 158 & 21 & 8 & 348 & 158 & 54 & 13 & 432 & 144 & 12 & 10 & 1197	& 500	& 96 &	 31\\
  pHPF(RVO+) & 141 & 48 & 4 & 1 & 313 & 175 & 23 & 8 & 344 & 175 & 50 & 26 & 457 & 155 & 5 & 0 & 1255 & 553 & 82 & 35  \\
  HPF(RVO+) & \textbf{184} & \textbf{79} & 6 & 4 & \textbf{384} & \textbf{211} & 24 & 10 & \textbf{477} & \textbf{245} & 42 & 21 & \textbf{479} & \textbf{169} & 3 & 0 & \textbf{1524}
  & \textbf{704} & 75 & 35 \\
 \hline
\end{tabular}

\end{footnotesize}
\label{tab:track_suc}
\end{table*}

\begin{figure*}
        \centering
		\makebox[\linewidth][c]{
        \begin{subfigure}[b]{0.24\textwidth}
                \centering
                \includegraphics[width=\textwidth]{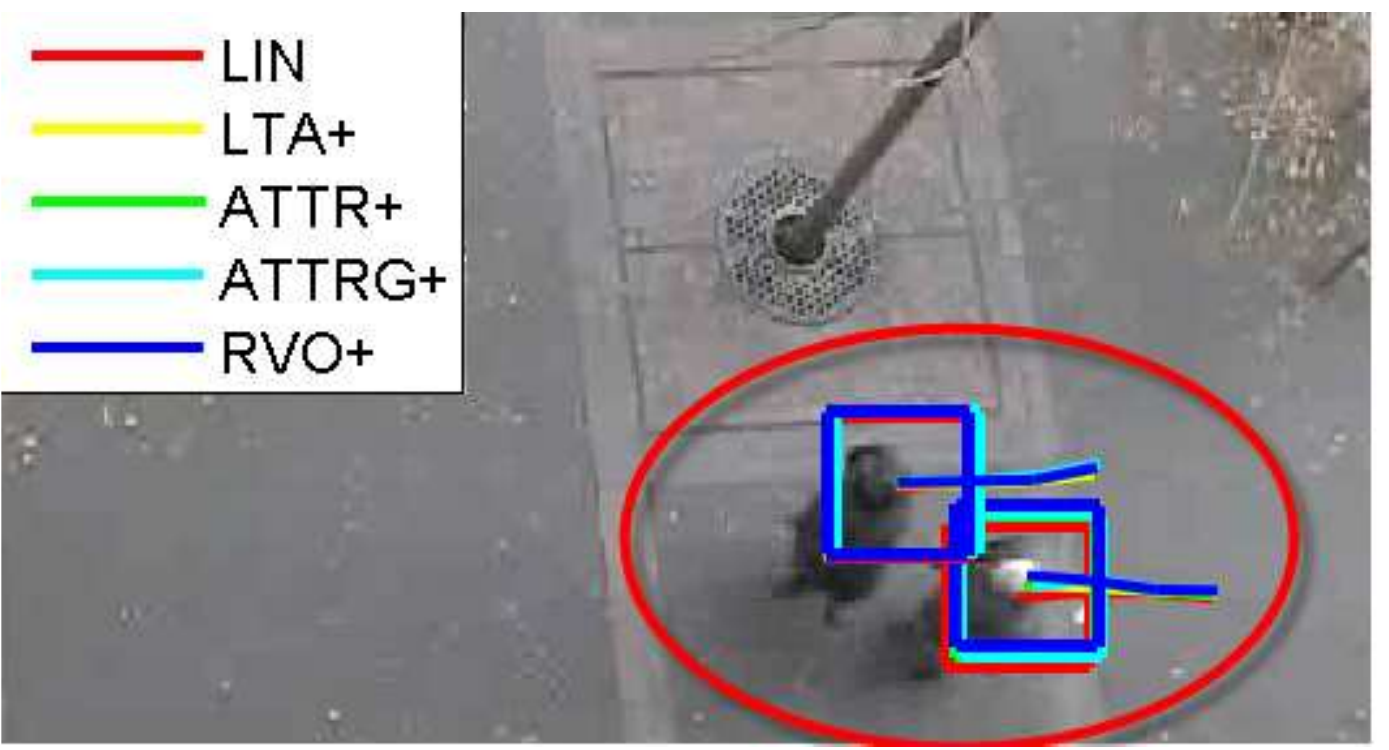}
        \end{subfigure}
        \hspace{-0.08 in}
        \begin{subfigure}[b]{0.24\textwidth}
                \centering
                \includegraphics[width=\textwidth]{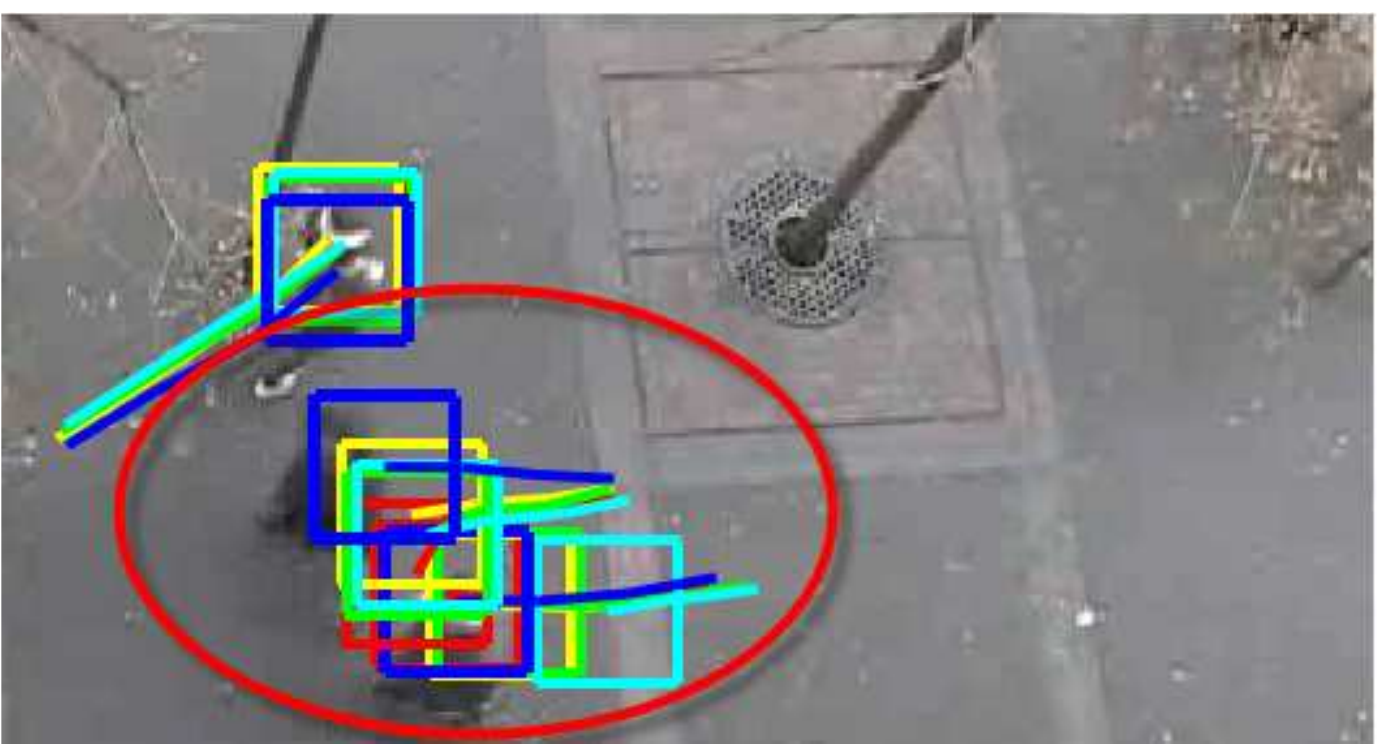}
        \end{subfigure}
        \hspace{0.05 in}
        \begin{subfigure}[b]{0.24\textwidth}
                \centering
                \includegraphics[width=\textwidth]{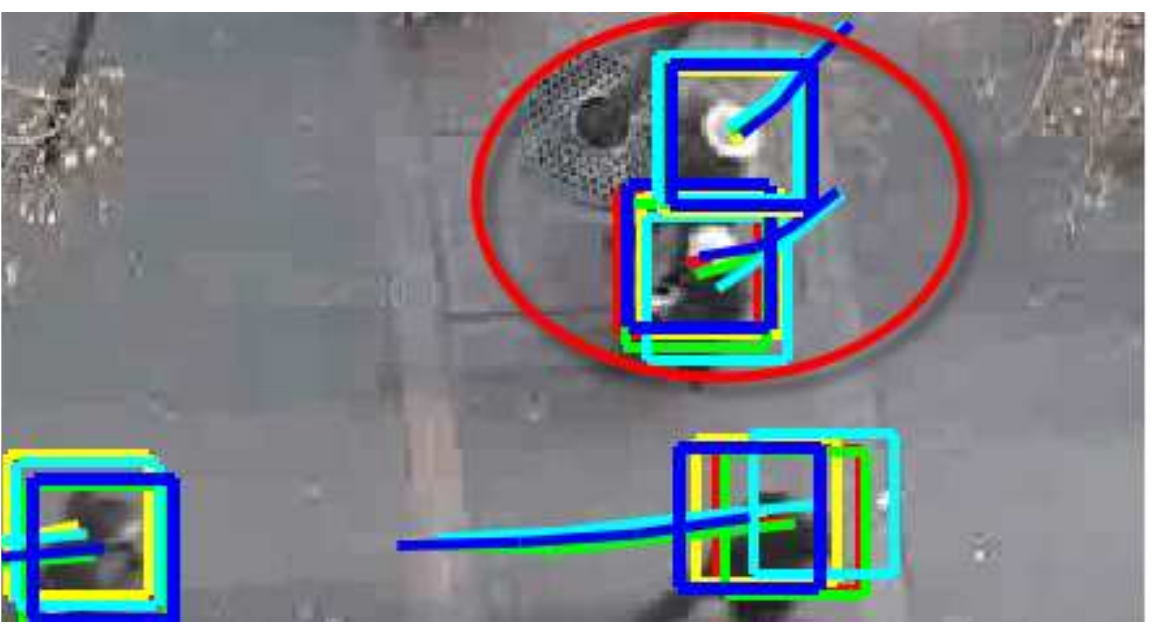}
        \end{subfigure}
        \hspace{-0.08 in}
        \begin{subfigure}[b]{0.24\textwidth}
                \centering
                \includegraphics[width=\textwidth]{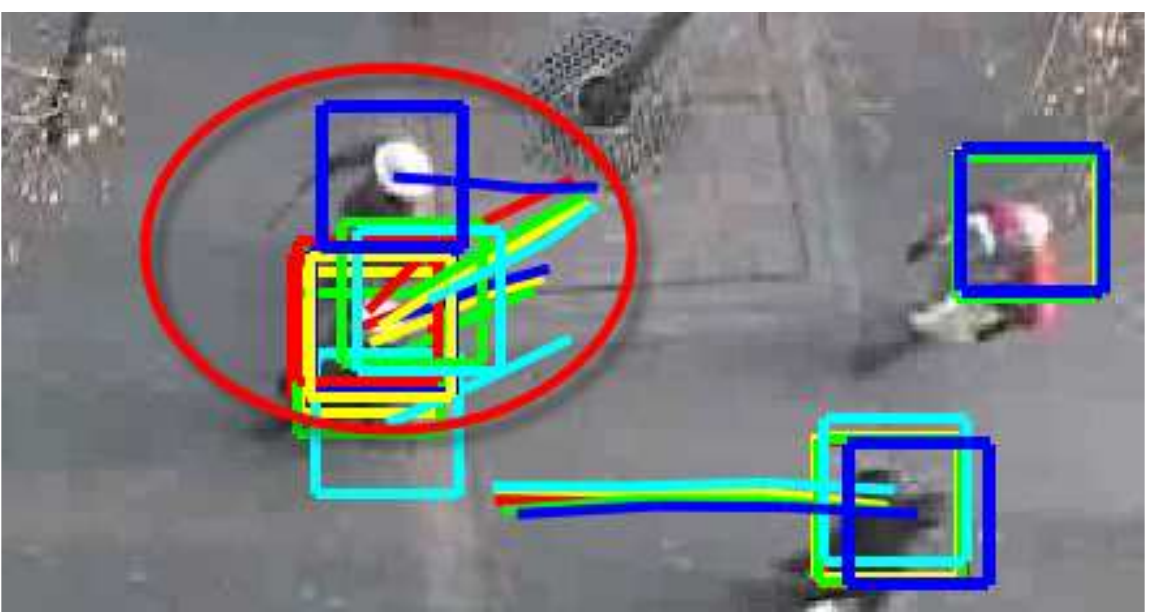}
        \end{subfigure}
		}\\
        \caption{{\small Two tracking examples (the first row and the second row) from \emph{Hotel}. Each example has two key frames with a circle highlighting that, as two pedestrians walking together, particles of one pedestrian are hijacked by the other pedestrian due to similar appearance, and hence the tracking boxes converge to one of the pedestrians. The motion prior provided by RVO+ keeps the tracking boxes on the right targets.}}
        \label{fig:track_hotel}
\end{figure*}

\begin{figure}
        \centering
        \begin{subfigure}[b]{0.38\textwidth}
                \centering
                \includegraphics[width=\textwidth]{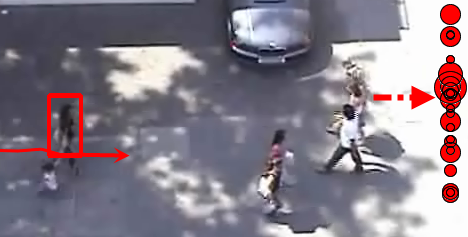}
        \end{subfigure}
        \begin{subfigure}[b]{0.38\textwidth}
                        \centering
                \includegraphics[width=\textwidth]{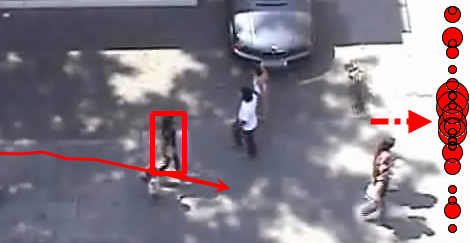}
        \end{subfigure}
        \caption{{\small Example of desired velocity.
          For the target pedestrian, the arrow indicates the velocity  of the person.
          On the right, the dot-line arrow is the mean desired velocity, and the red circles are the weighted particles (the size indicates the importance), representing desired velocities projected to destinations.
          When the person deviates to avoid collision with oncoming pedestrians, the desired velocity remains almost the same.}}
  		\label{fig:dest_est}
\end{figure}

\textbf{Tracking results.}
The tracking results are presented in Table~\ref{tab:track_suc}.
Compared to other motion models,
RVO+ has the highest numbers of successful tracks in all the sequences.
In addition, estimating the desired velocity increases the overall number of successful tracks for all motion \mbox{models}.
When learning the desired velocity, the proportion of ``long'' successful tracks increases compared to the proportion of ``short'' STs (e.g., for ATTR/ATTR+, the short STs increase by 14\%, whereas the long STs increase by 30\%).  These results suggest that online learning is useful in terms of long-term tracking. Online-learned desired velocities can prevent the trackers from drifting away. As shown in Figure~\ref{fig:dest_est}, the desired velocity remains stable as the tracked person steers away to avoid collision.
%
Note that on \emph{Zara02} and \emph{Student}, RVO+ has large number of IDSs. This is because the desired velocity increases the variation of the particle distribution to prevent losing track, but as a consequence reduces the tracking precision in crowded scenes. The number of IDSs are reduced using our HPF model.
Figure~\ref{fig:track_hotel} shows two tracking examples on \emph{Hotel} using RVO+ and other models.  
When the two targets are near to each other and have similar appearances, particles of one target are easily ``hijacked" by the other target.  Here, the reciprocal collision avoidance of RVO+ prevents hijacking.

\begin{figure}
        \hspace*{-0.2 in}
        \centering
        \begin{subfigure}[b]{0.38\textwidth}
        	\centering
			\includegraphics[width=\textwidth]{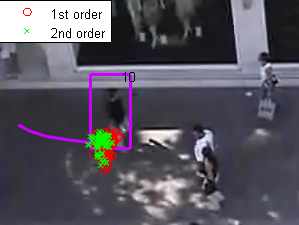}
        \end{subfigure}
        \begin{subfigure}[b]{0.38\textwidth}
        	\centering
            \includegraphics[width=\textwidth]{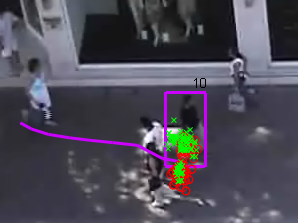}
        \end{subfigure}

        \caption{{\small HPF example on  \emph{Zara01}.
        The red circles represent the 1st-order particle distribution and the green crosses represent 2nd-order particle distribution. 
        When the target pedestrian bypass the oncoming couple, a large portion of 1st-order particles deviate due to the occlusion whereas the majority of 2nd-order particles maintain the tracking boxes on the right target.}}
        \label{fig:hpf_track_cmp}
\end{figure}

\begin{figure*}
        \centering
		\makebox[\linewidth][c]{
        \begin{subfigure}[b]{0.28\textwidth}
                \centering
                \includegraphics[width=\textwidth]{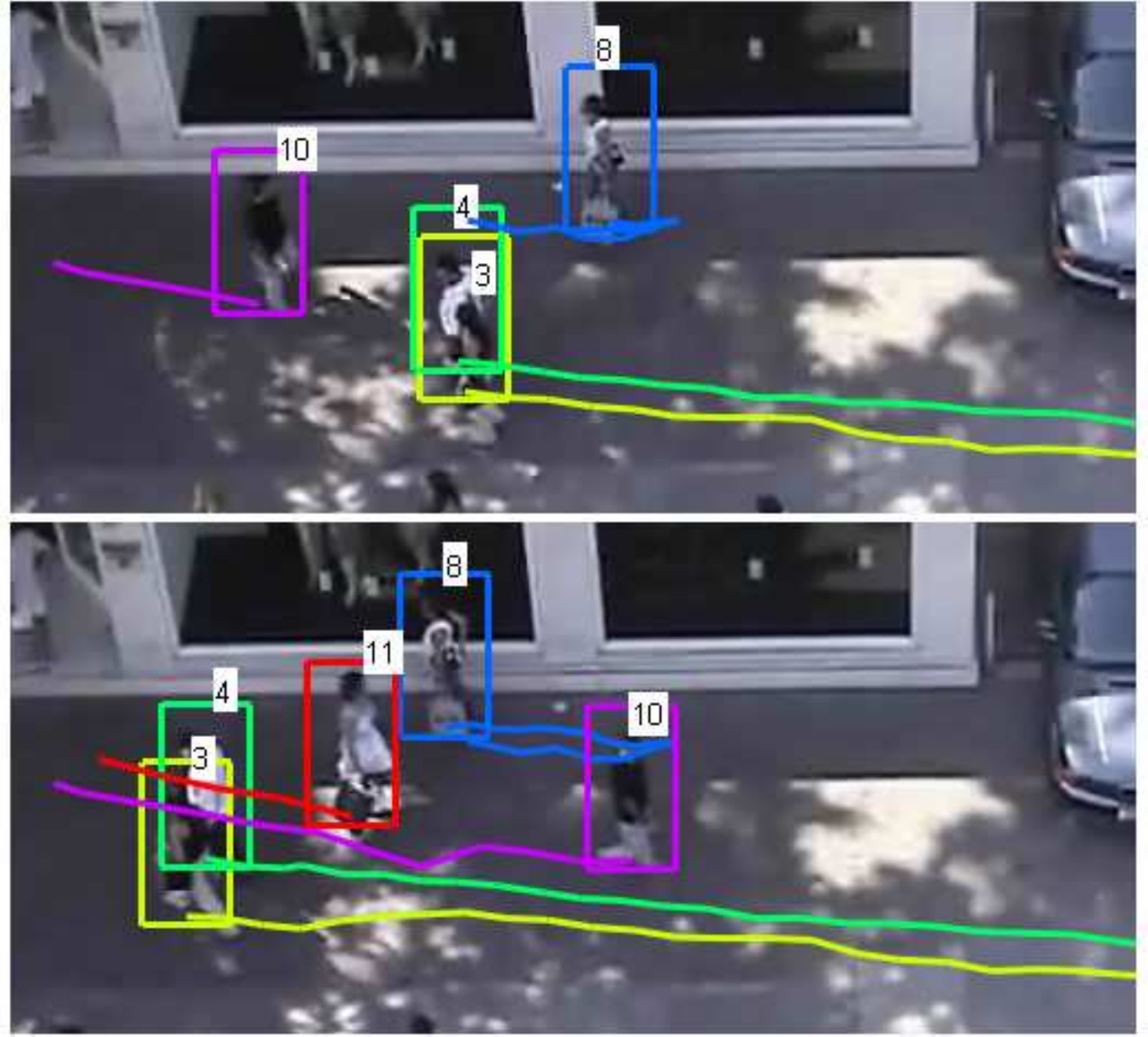}				
                \caption{}
        \end{subfigure}
        \begin{subfigure}[b]{0.24\textwidth}
                \centering
                \includegraphics[width=\textwidth]{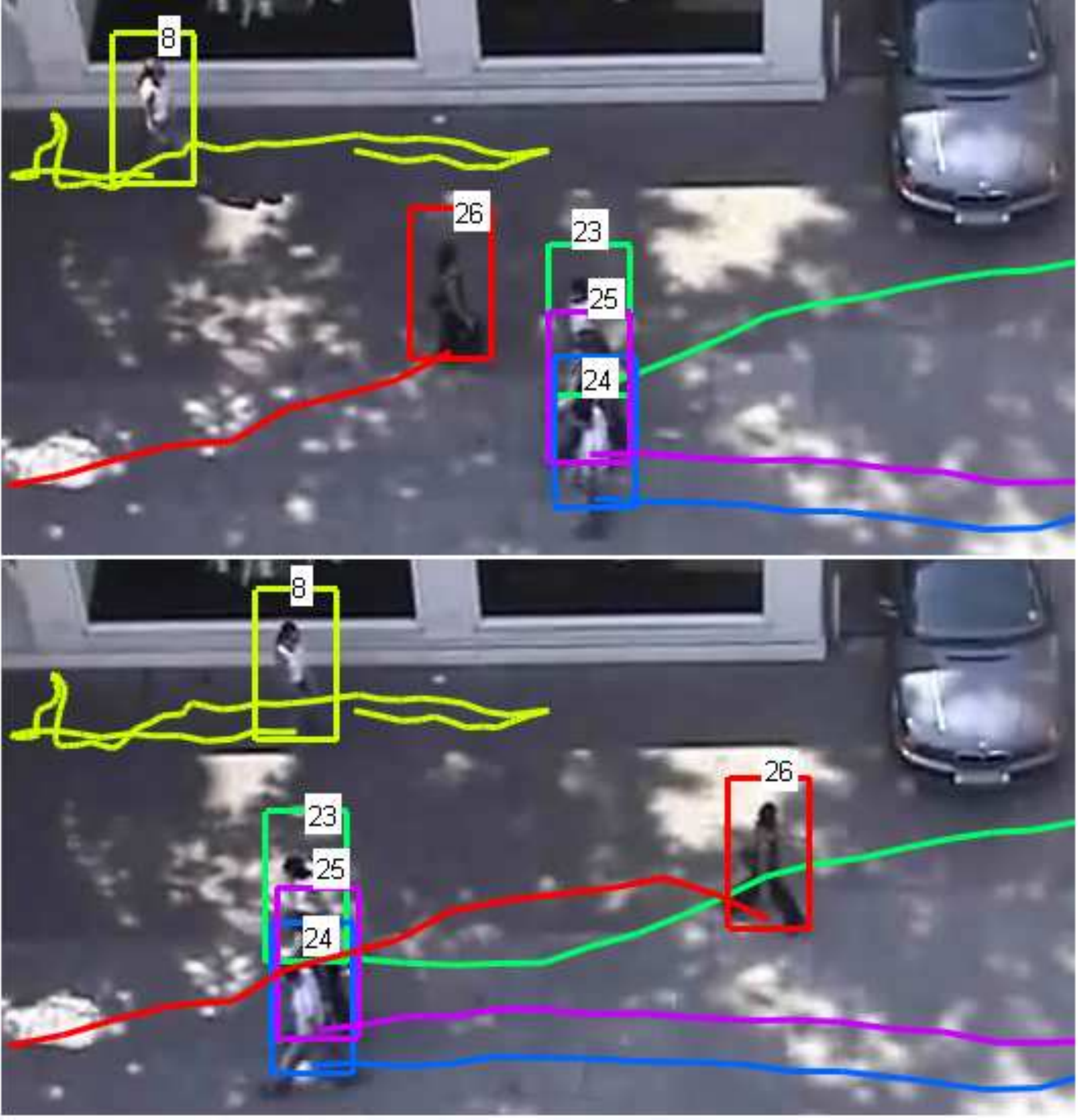}
                \caption{}
        \end{subfigure}
        \begin{subfigure}[b]{0.25\textwidth}
                \centering
                \includegraphics[width=\textwidth]{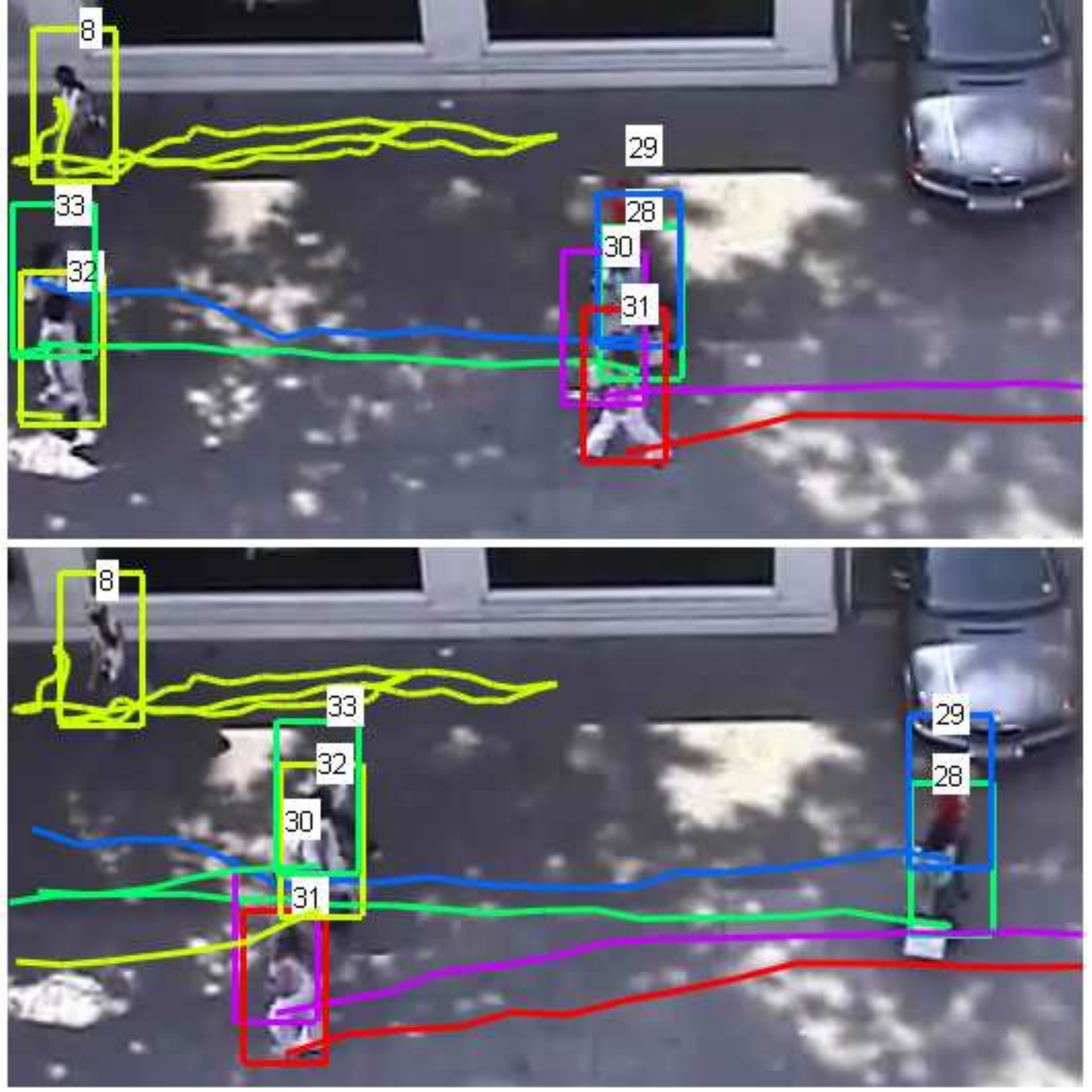}
                \caption{}
        \end{subfigure}
        \begin{subfigure}[b]{0.25\textwidth}
                \centering
                \includegraphics[width=\textwidth]{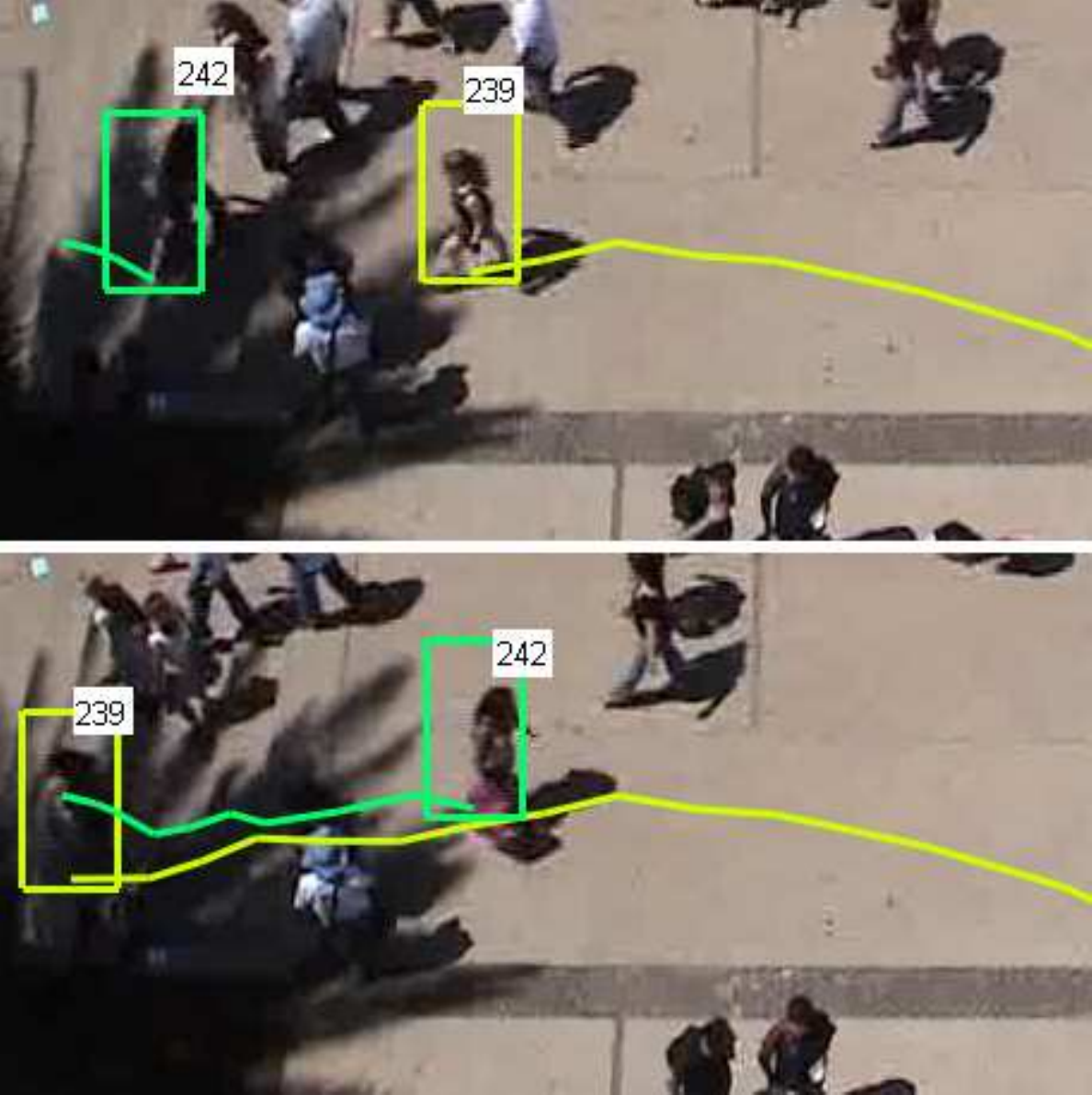}
                \caption{}
        \end{subfigure}
        }\\
        \caption{{\small Four examples from \emph{Zara01}/\emph{Student} using HPF for tracking pedestrians. Each example shows the trajectories of tracking in two frames. The first 3 examples include pedestrian pairs tracked  through interaction and occlusions. There is one person (ID=8) lingering around with irregular trajectories and frequent occlusions, and HPF still tracks her throughout the video. The 4th example shows the benefits of our motion model 
        when the appearance model fails; 
       although the interaction takes place in the shadow,
        HPF maintains the track.}}
        \vspace{-0.1in}
        \label{fig:track_zara}
\end{figure*}

\begin{figure*}
        \centering
        \makebox[\linewidth][c]{
        \begin{subfigure}[b]{0.32\textwidth}
                \centering
                \includegraphics[width=\textwidth]{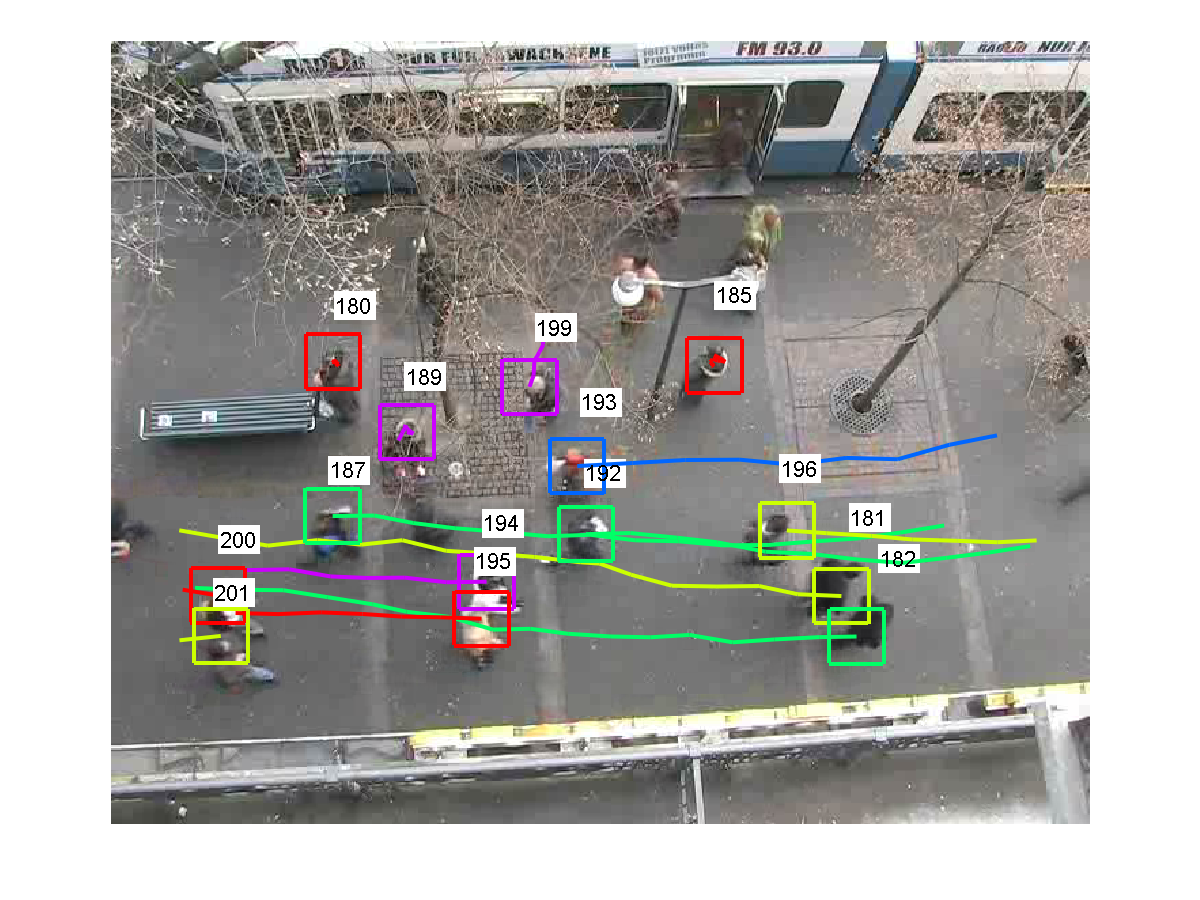}
        \end{subfigure}
        \hspace{-0.3 in}
        \begin{subfigure}[b]{0.32\textwidth}
                \centering
                \includegraphics[width=\textwidth]{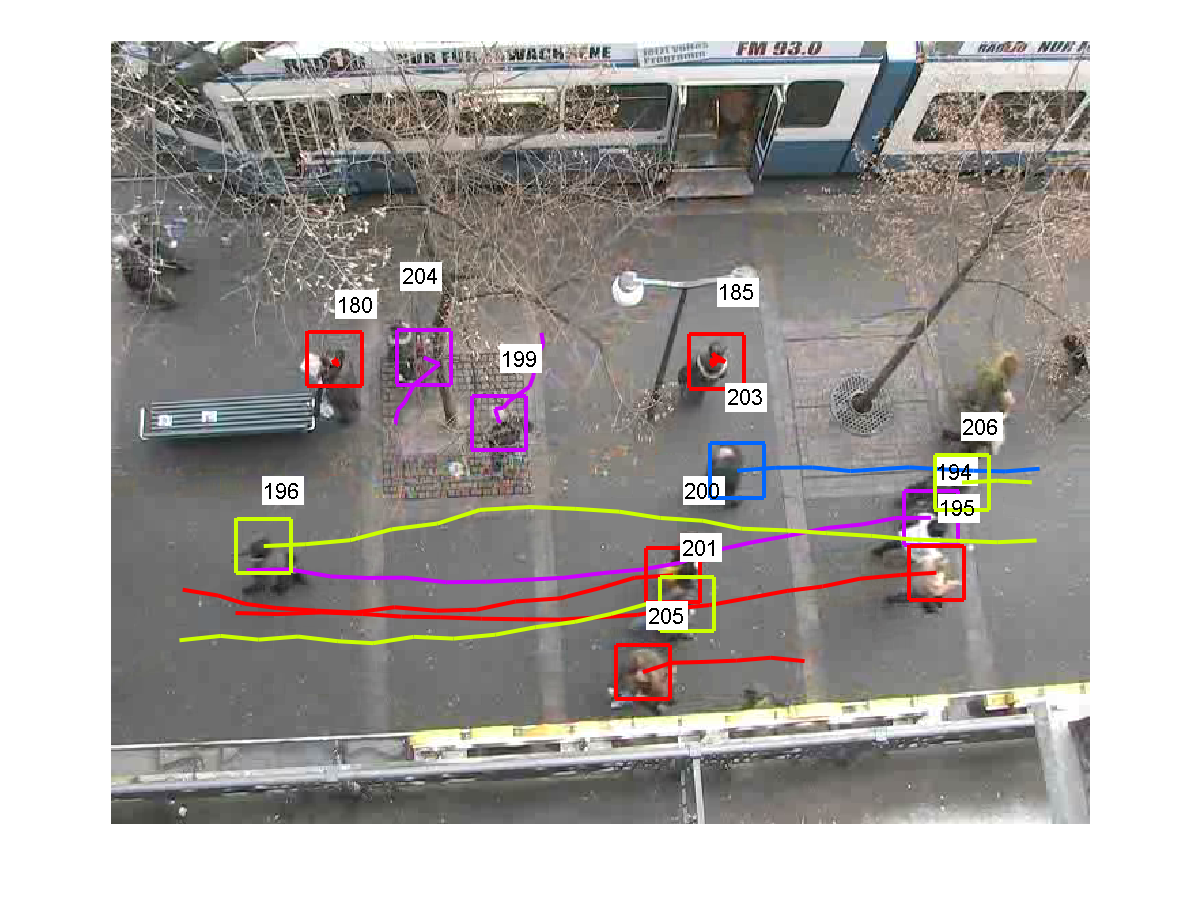}
        \end{subfigure}
        \hspace{-0.3 in}
        \begin{subfigure}[b]{0.32\textwidth}
                \centering
                \includegraphics[width=\textwidth]{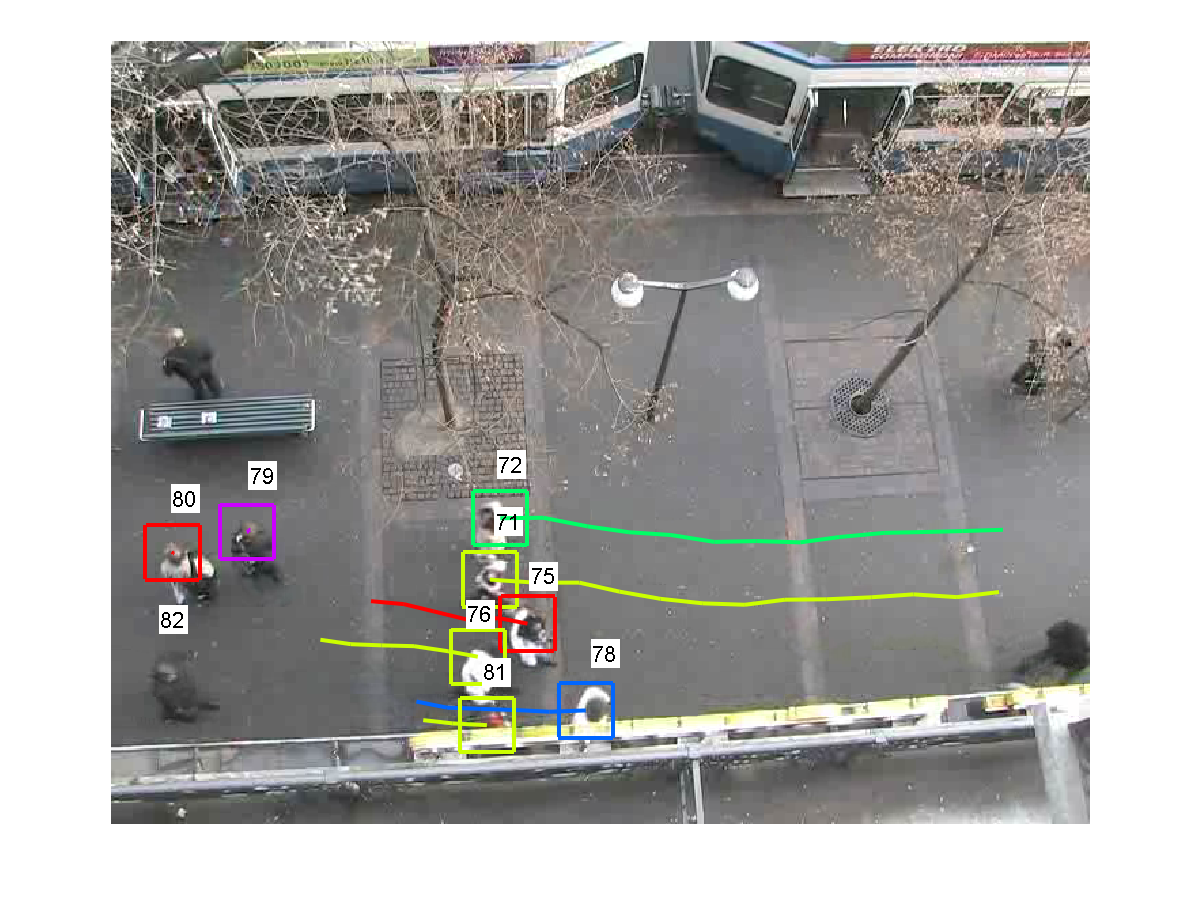}
        \end{subfigure}
        \hspace{-0.3 in}
        \begin{subfigure}[b]{0.32\textwidth}
                \centering
                \includegraphics[width=\textwidth]{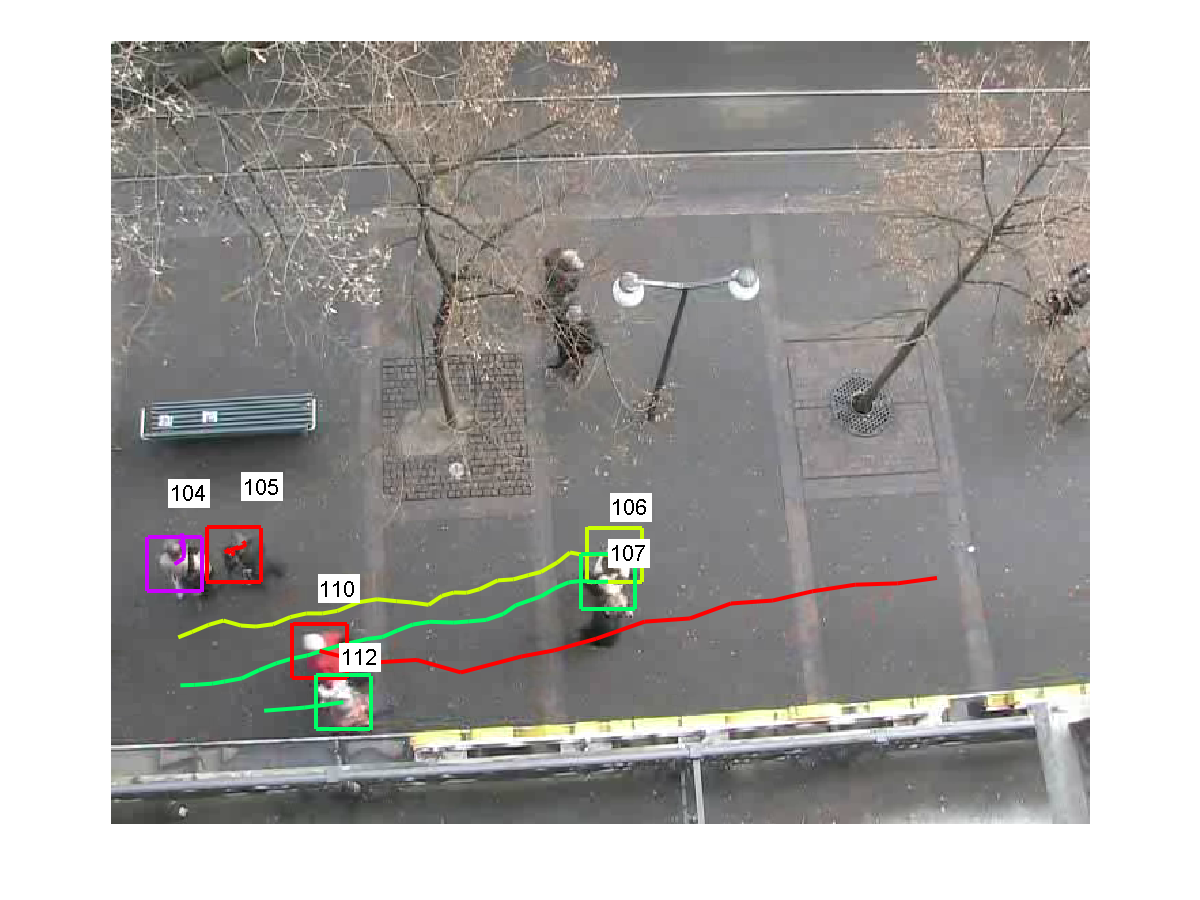}
        \end{subfigure}
        }\\	
        \centering
		\makebox[\linewidth][c]{
        \begin{subfigure}[b]{0.32\textwidth}
                \centering
                \includegraphics[width=\textwidth]{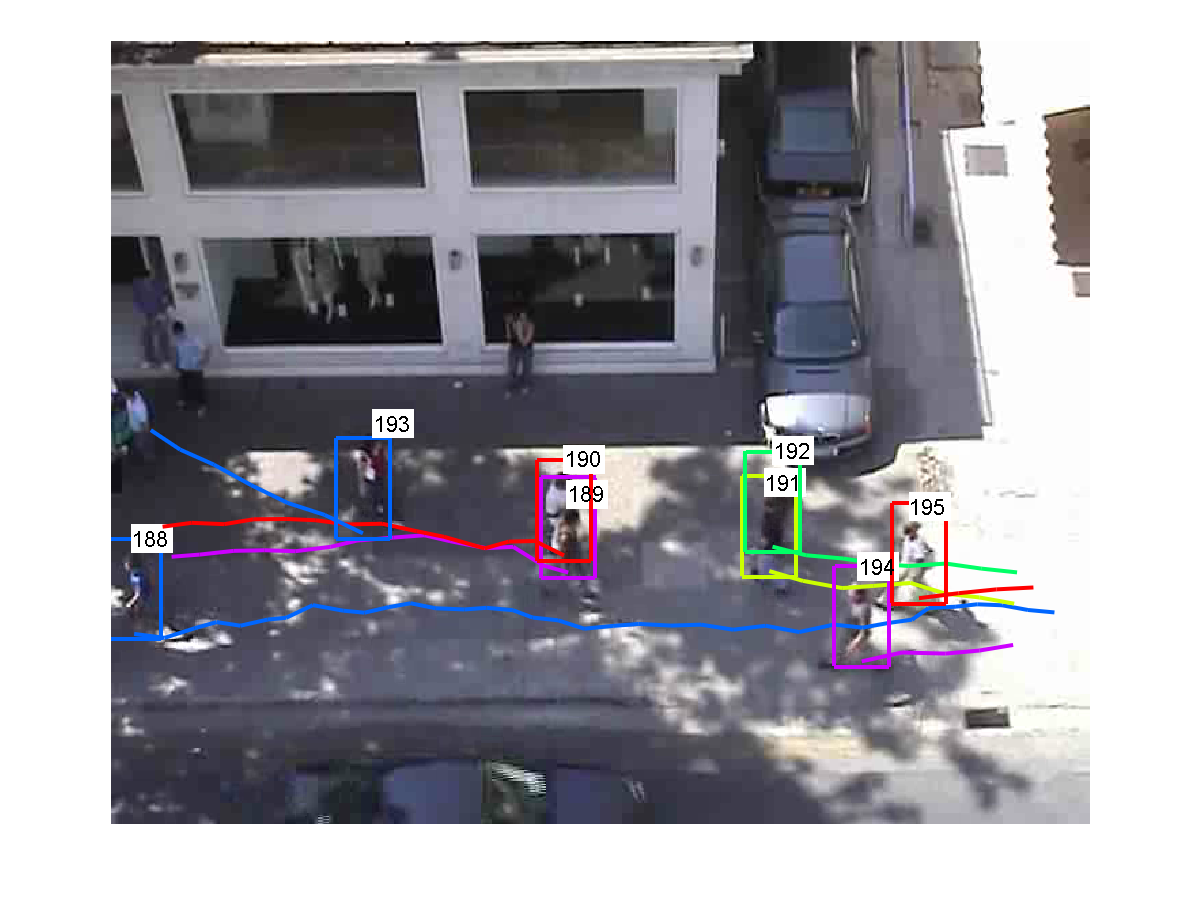}
        \end{subfigure}
        \hspace{-0.3 in}
        \begin{subfigure}[b]{0.32\textwidth}
                \centering
                \includegraphics[width=\textwidth]{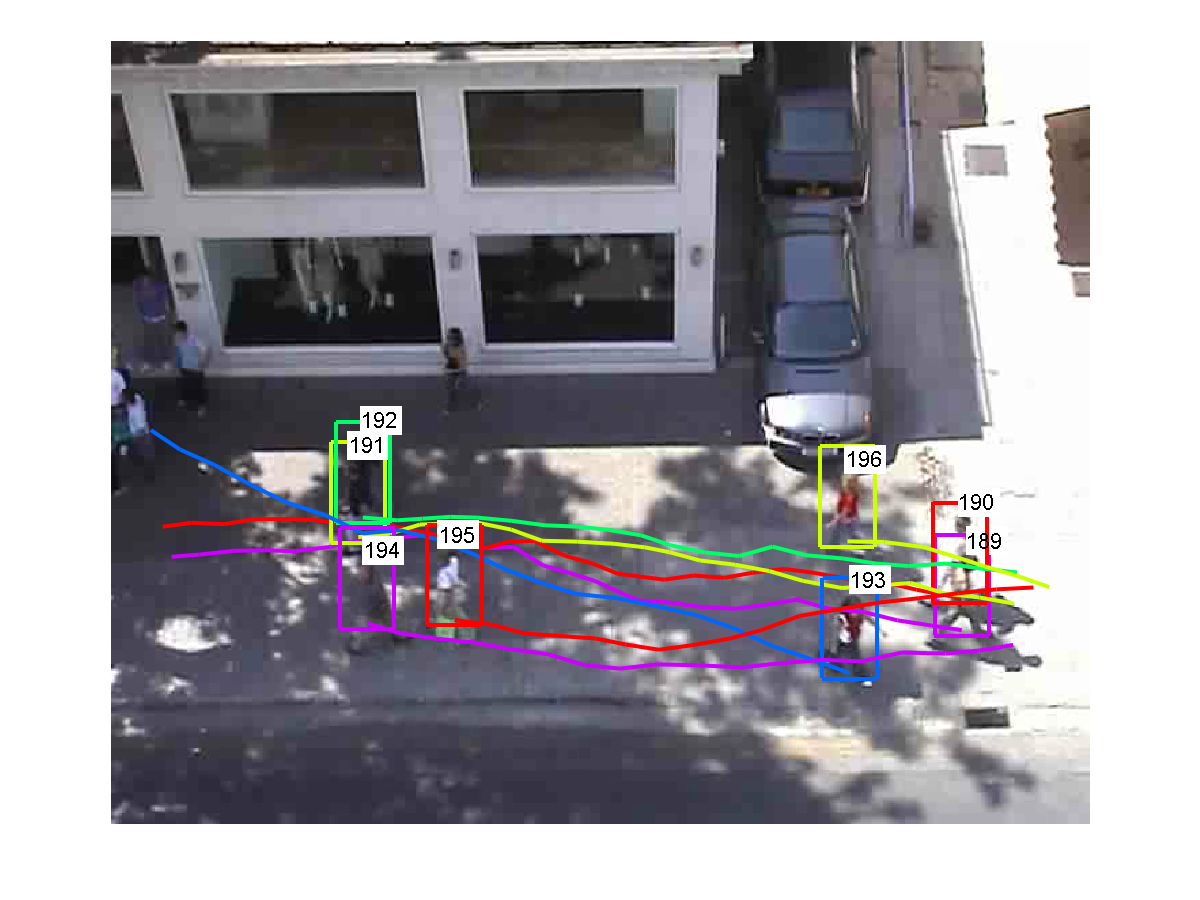}
        \end{subfigure}
        \hspace{-0.3 in}
        \begin{subfigure}[b]{0.32\textwidth}
                \centering
                \includegraphics[width=\textwidth]{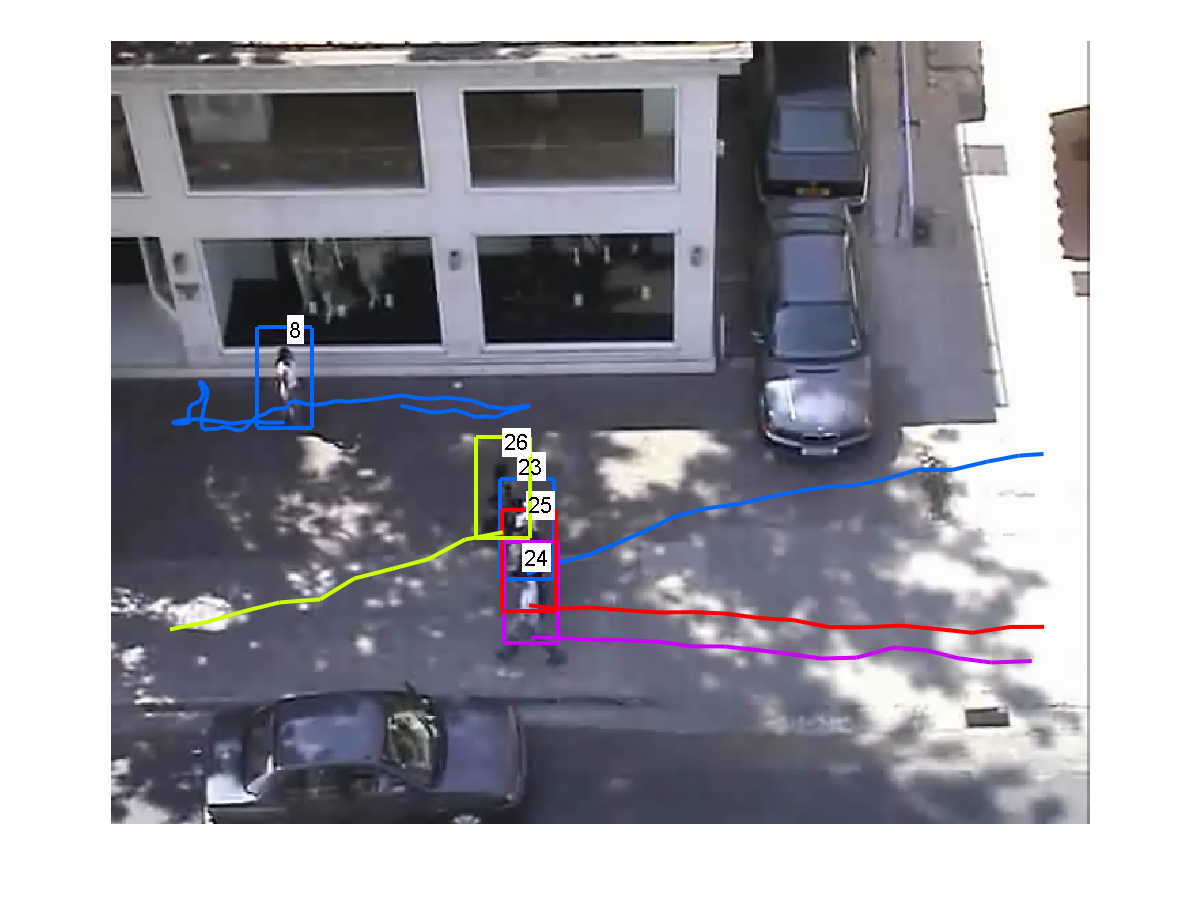}
        \end{subfigure}
        \hspace{-0.3 in}
        \begin{subfigure}[b]{0.32\textwidth}
                \centering
                \includegraphics[width=\textwidth]{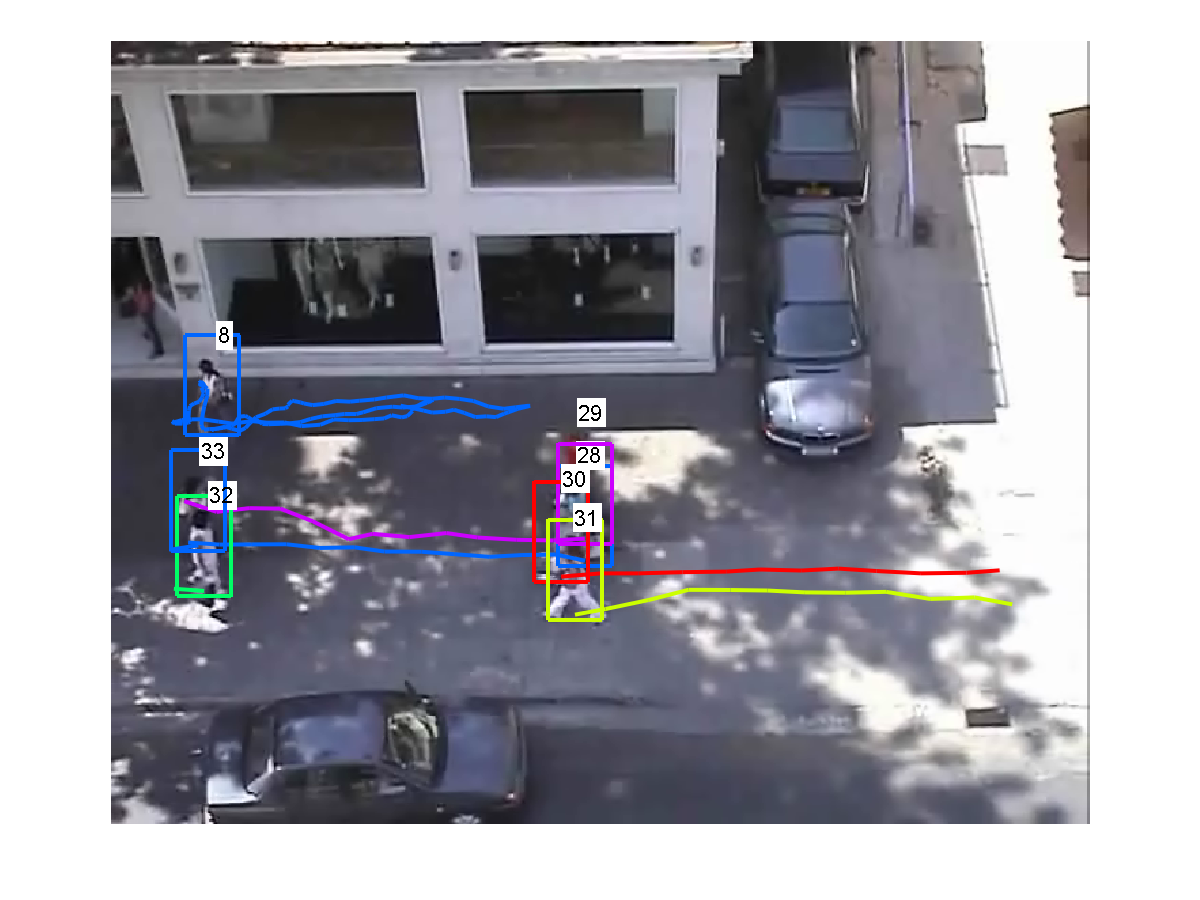}
        \end{subfigure}
        }\\
		\vspace{-0.1in}
        \caption{{\small Eight tracking examples on \emph{Hotel}, \emph{Zara01} and \emph{Zara02}. HPF keeps track of pedestrians well with the interactions and occlusions among pedestrians.}}
        \label{fig:track_zara_add}
        \vspace{-0.15in}
\end{figure*}

\begin{figure*}
        \centering
		\makebox[\linewidth][c]{
        \begin{subfigure}[b]{0.32\textwidth}
                \centering
                \includegraphics[width=\textwidth]{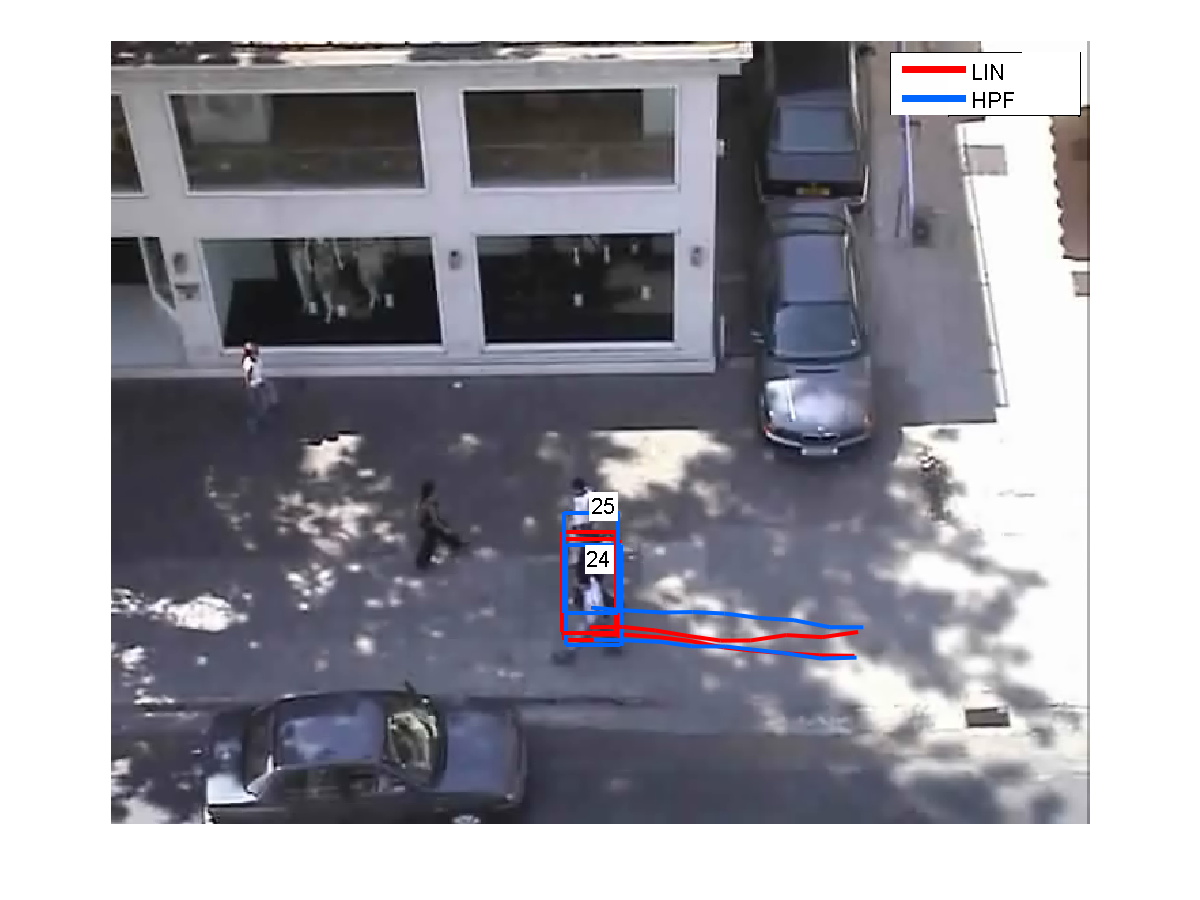}
        \end{subfigure}
        \hspace{-0.3 in}
        \begin{subfigure}[b]{0.32\textwidth}
                \centering
                \includegraphics[width=\textwidth]{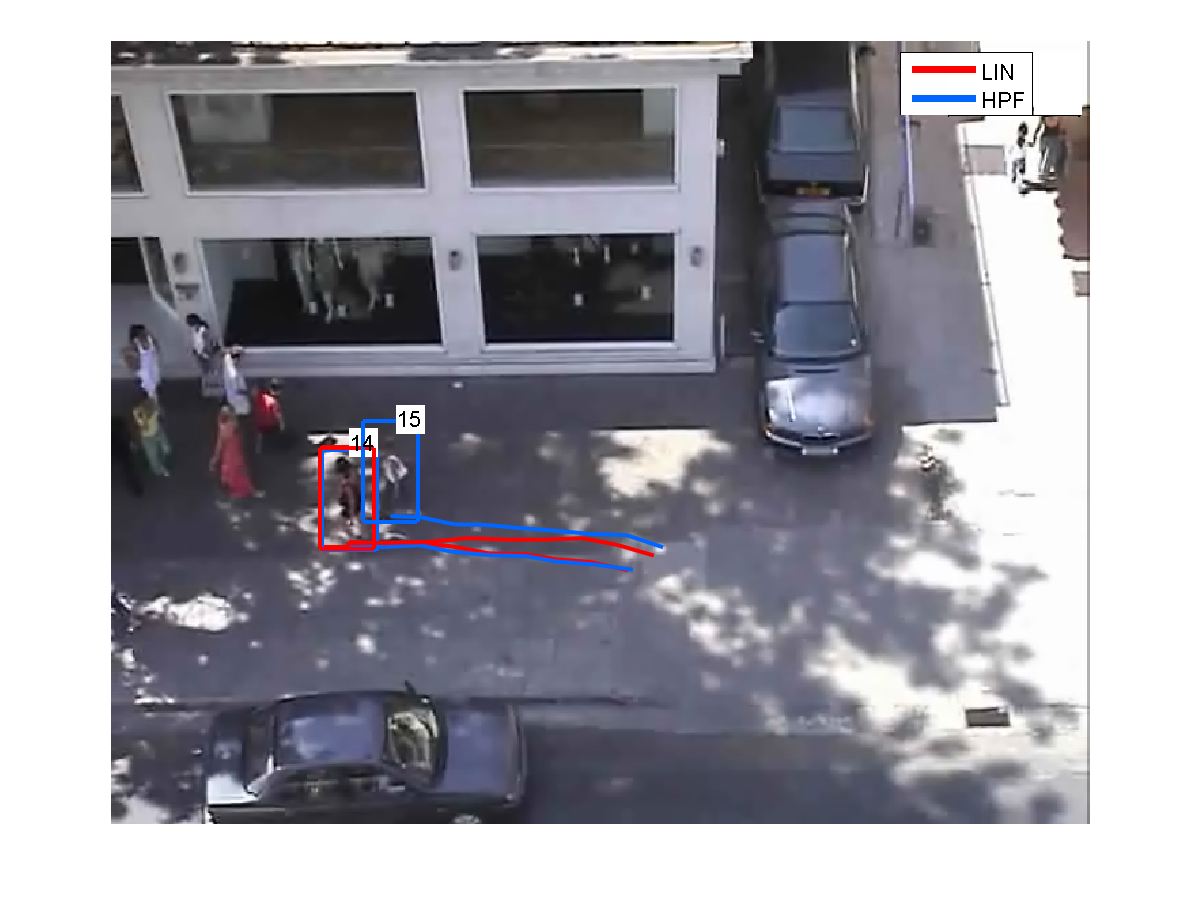}
        \end{subfigure}
        \hspace{-0.3 in}
        \begin{subfigure}[b]{0.32\textwidth}
                \centering
                \includegraphics[width=\textwidth]{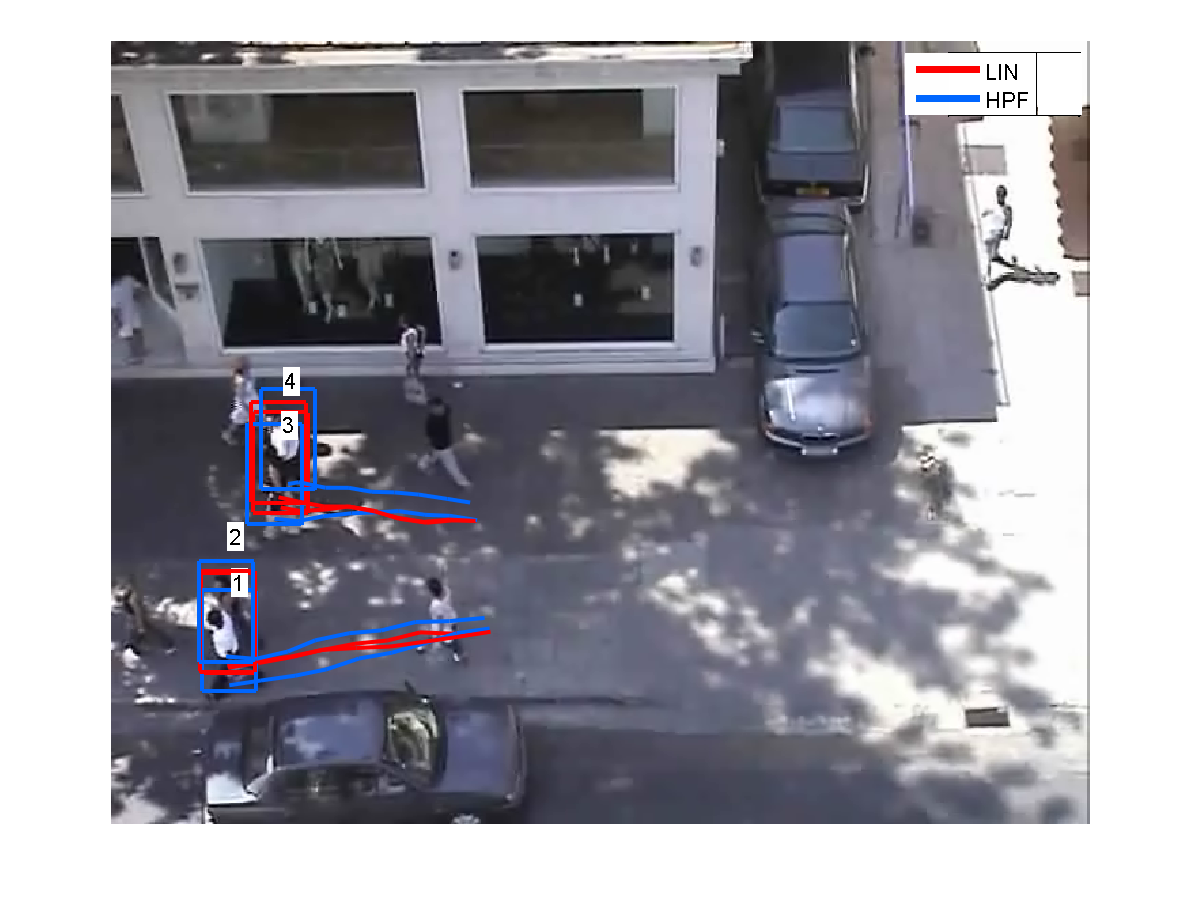}
        \end{subfigure}
        \hspace{-0.3 in}
        \begin{subfigure}[b]{0.32\textwidth}
                \centering
                \includegraphics[width=\textwidth]{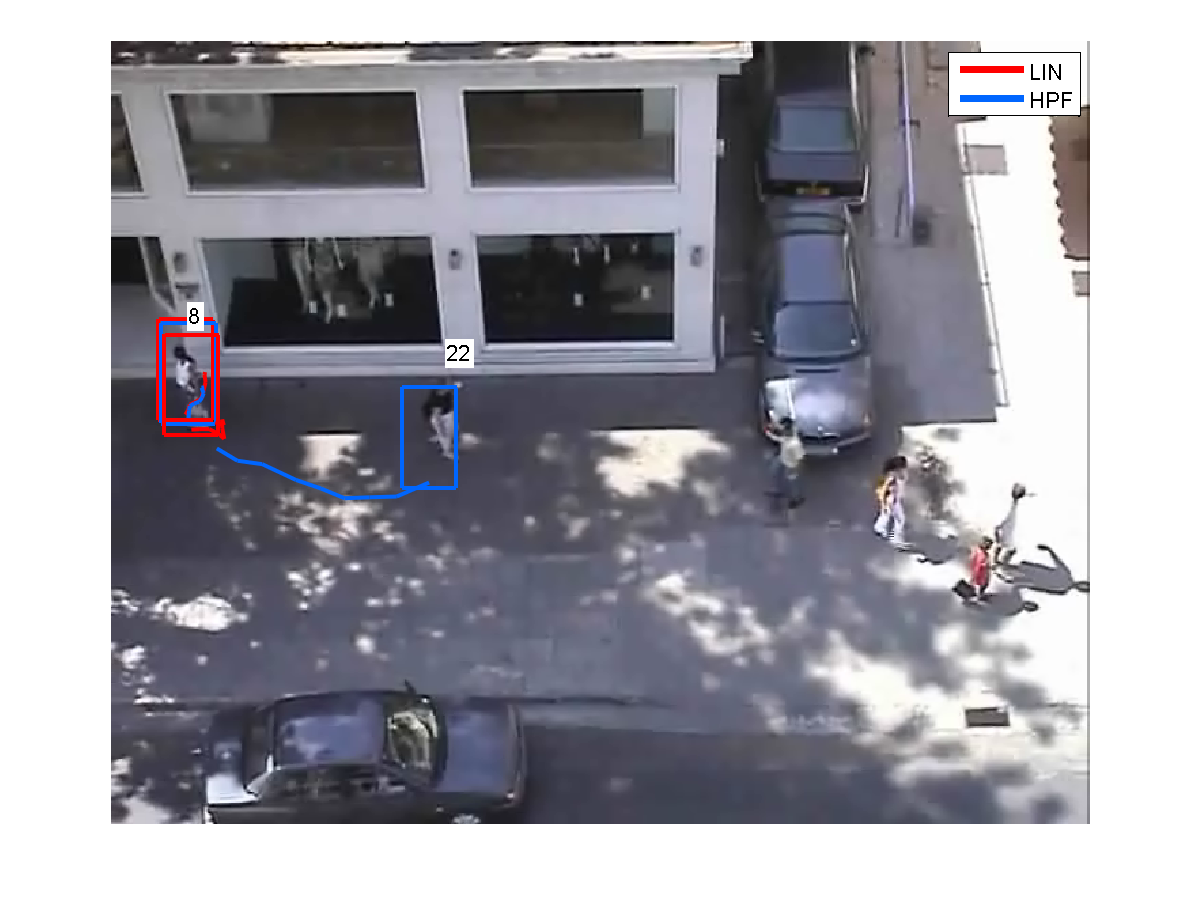}
        \end{subfigure}
        }\\
		\vspace{-0.1in}
        \caption{{\small Four tracking examples on \emph{Zara01} comparing LIN and HPF. As the targets are close to each other, LIN has ID switches (one's particles are `hijacked' by the other), while HPF does not have this problem.}}
        \label{fig:ids_zara}
        \vspace{-0.15in}
\end{figure*}

\begin{figure*}
        \centering
		\makebox[\linewidth][c]{
        \begin{subfigure}[b]{0.32\textwidth}
                \centering
                \includegraphics[width=\textwidth]{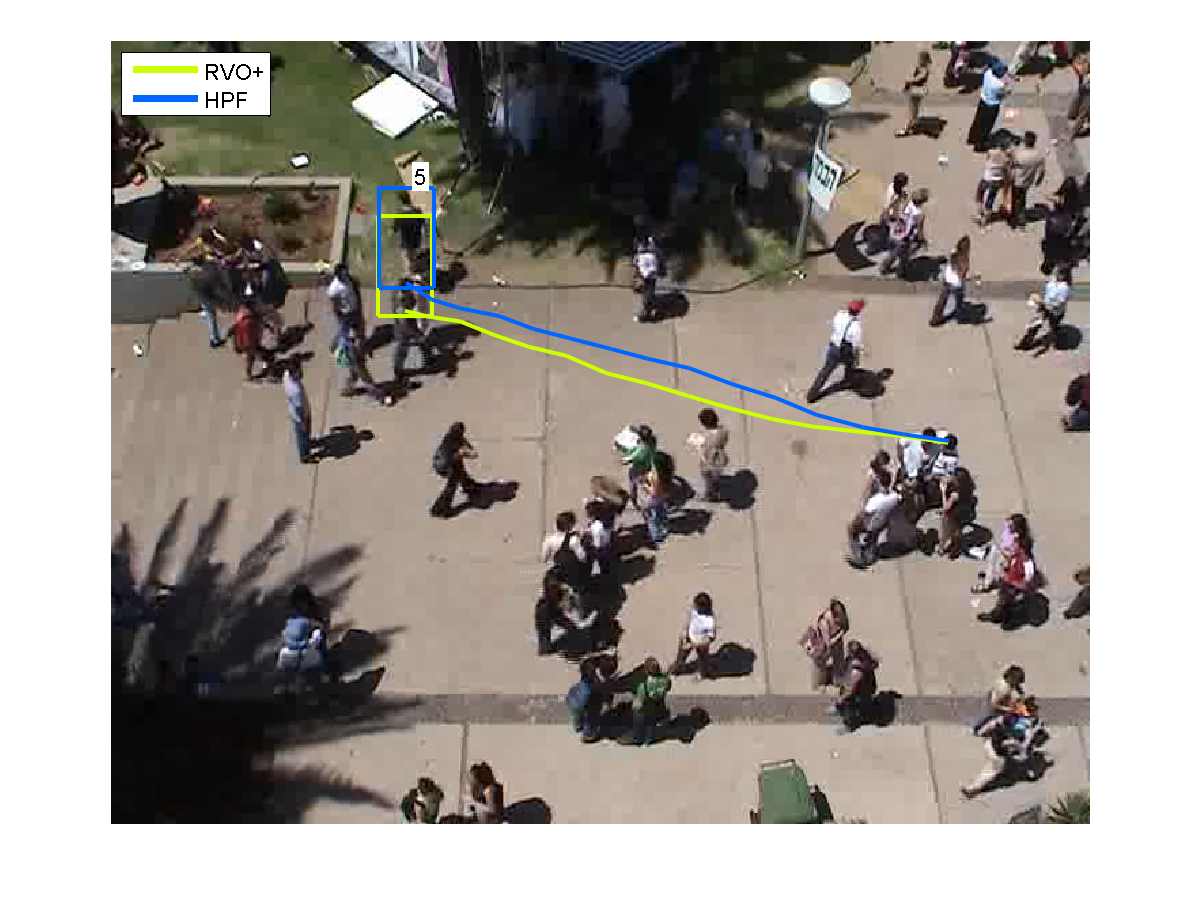}
        \end{subfigure}
        \hspace{-0.3 in}
        \begin{subfigure}[b]{0.32\textwidth}
                \centering
                \includegraphics[width=\textwidth]{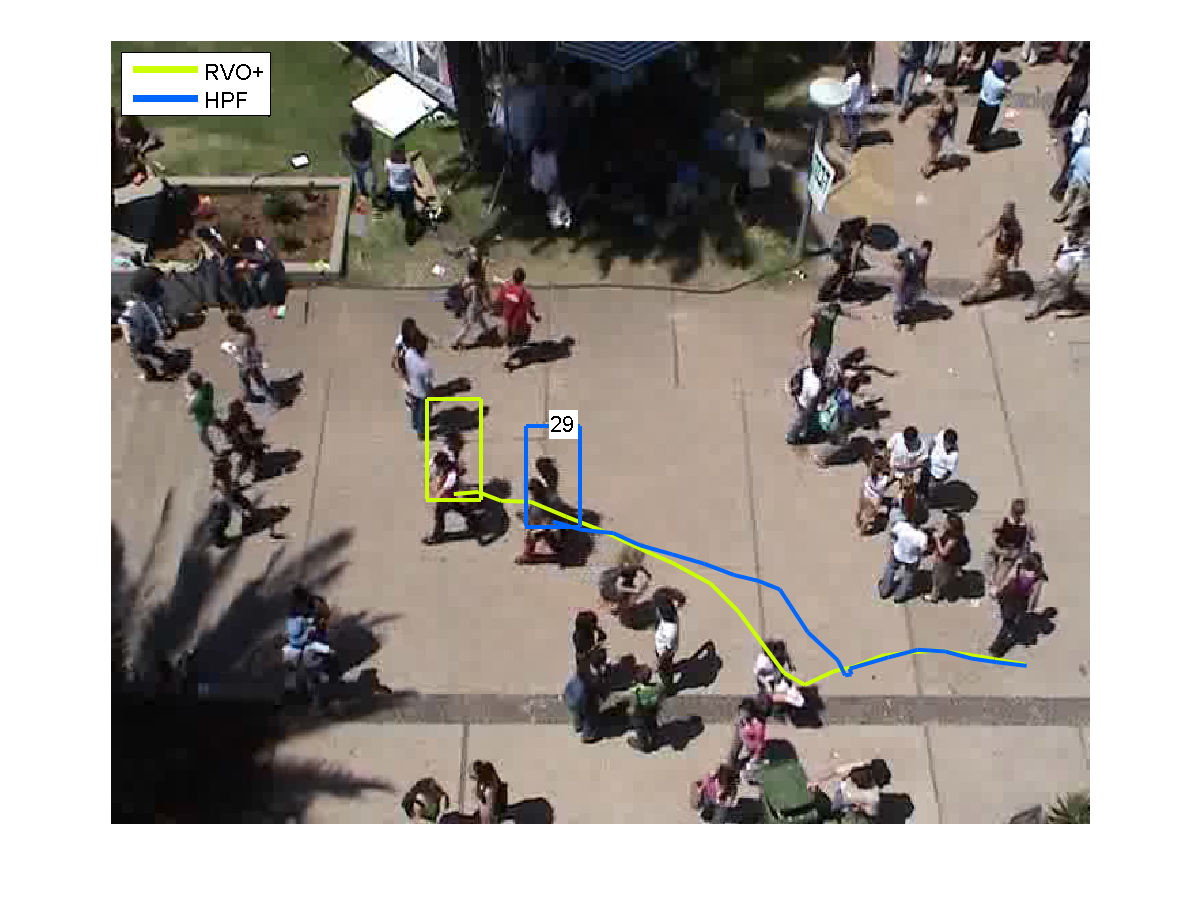}
        \end{subfigure}
        \hspace{-0.3 in}
        \begin{subfigure}[b]{0.32\textwidth}
                \centering
                \includegraphics[width=\textwidth]{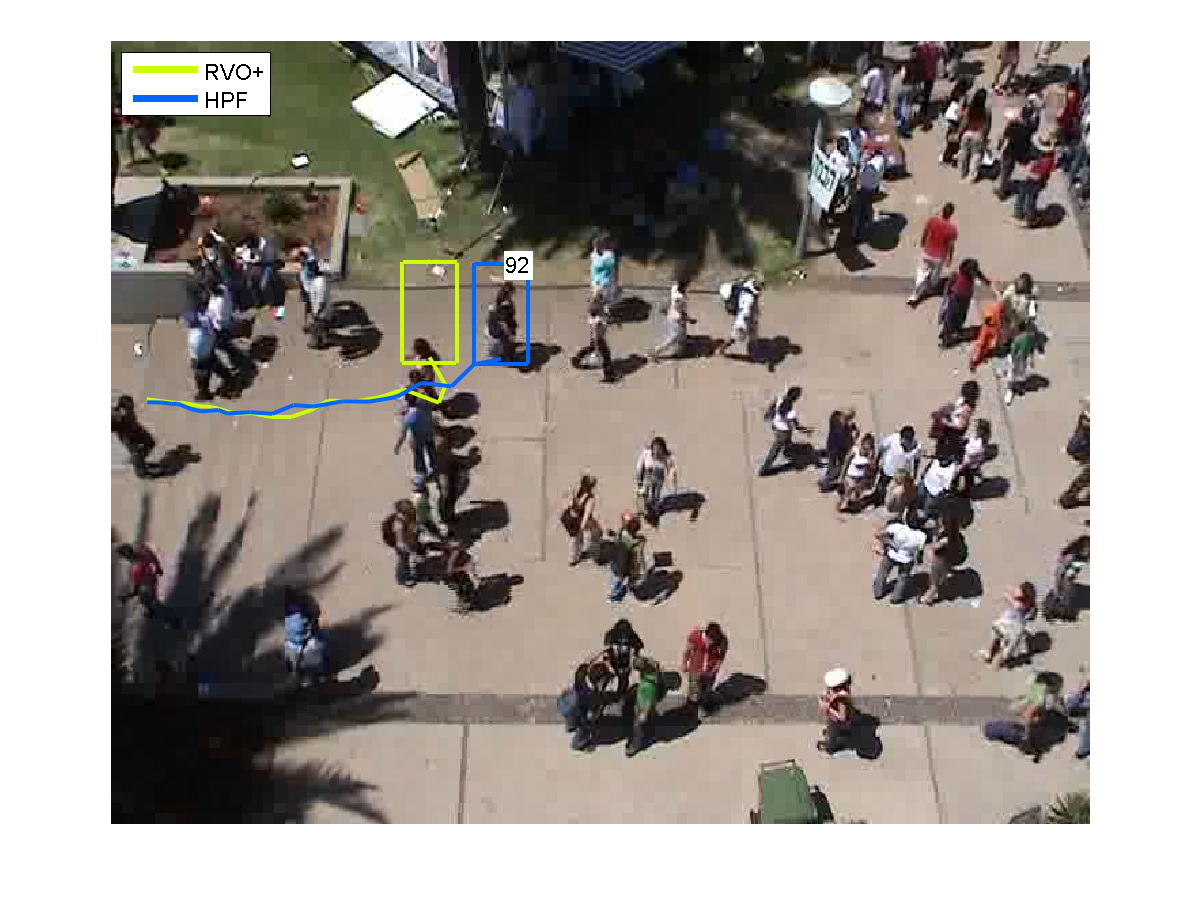}
        \end{subfigure}
        \hspace{-0.3 in}
        \begin{subfigure}[b]{0.32\textwidth}
                \centering
                \includegraphics[width=\textwidth]{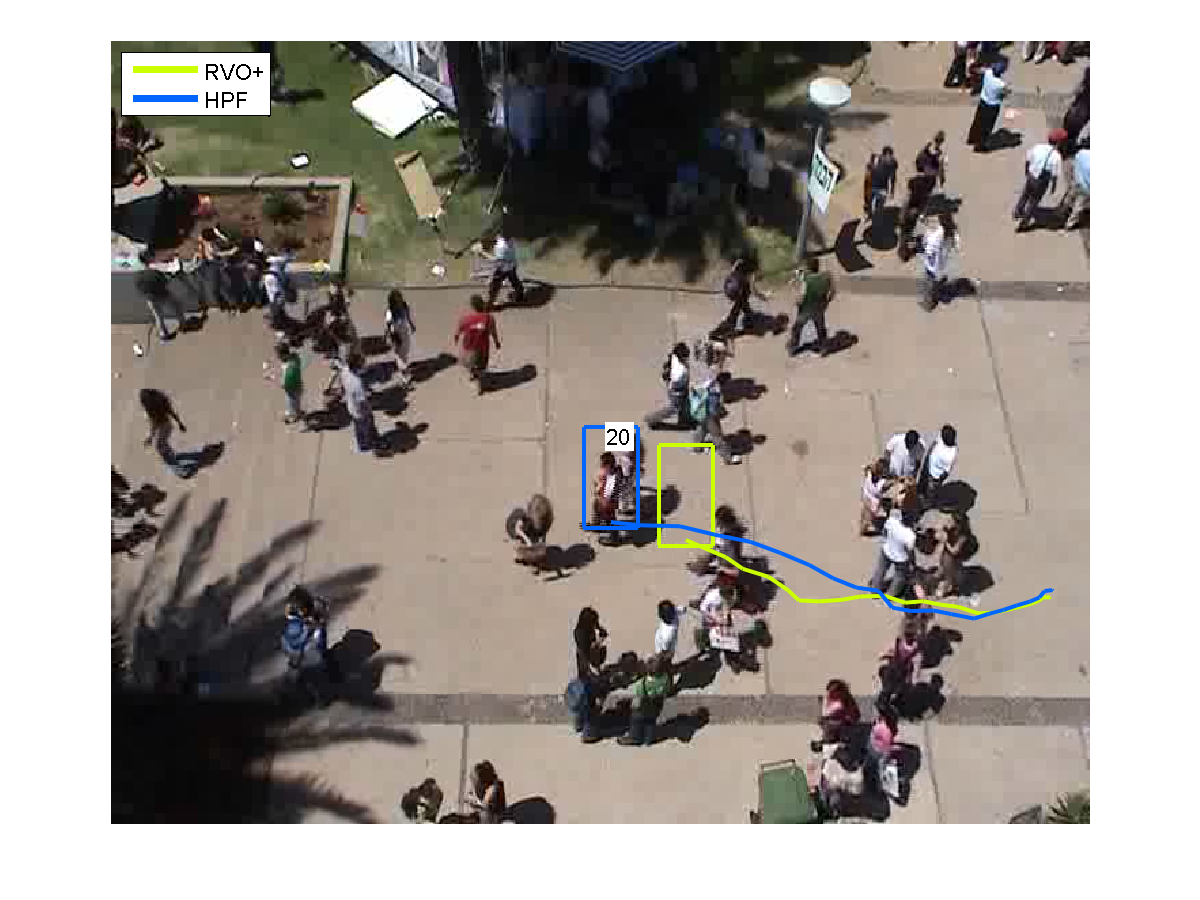}
        \end{subfigure}
        }\\
		\vspace{-0.1in}
        \caption{{\small Four representative tracking examples on \emph{Student} comparing RVO+ and HPF. As pedestrians are occluded, RVO+ drifts away or has ID switches. In constrast, HPF performs better using higher-order particles. }}
        \label{fig:ids_stud}
        \vspace{-0.15in}
\end{figure*}

\begin{figure*}
        \centering
		\makebox[\linewidth][c]{
        \begin{subfigure}[b]{0.32\textwidth}
                \centering
                \includegraphics[width=\textwidth]{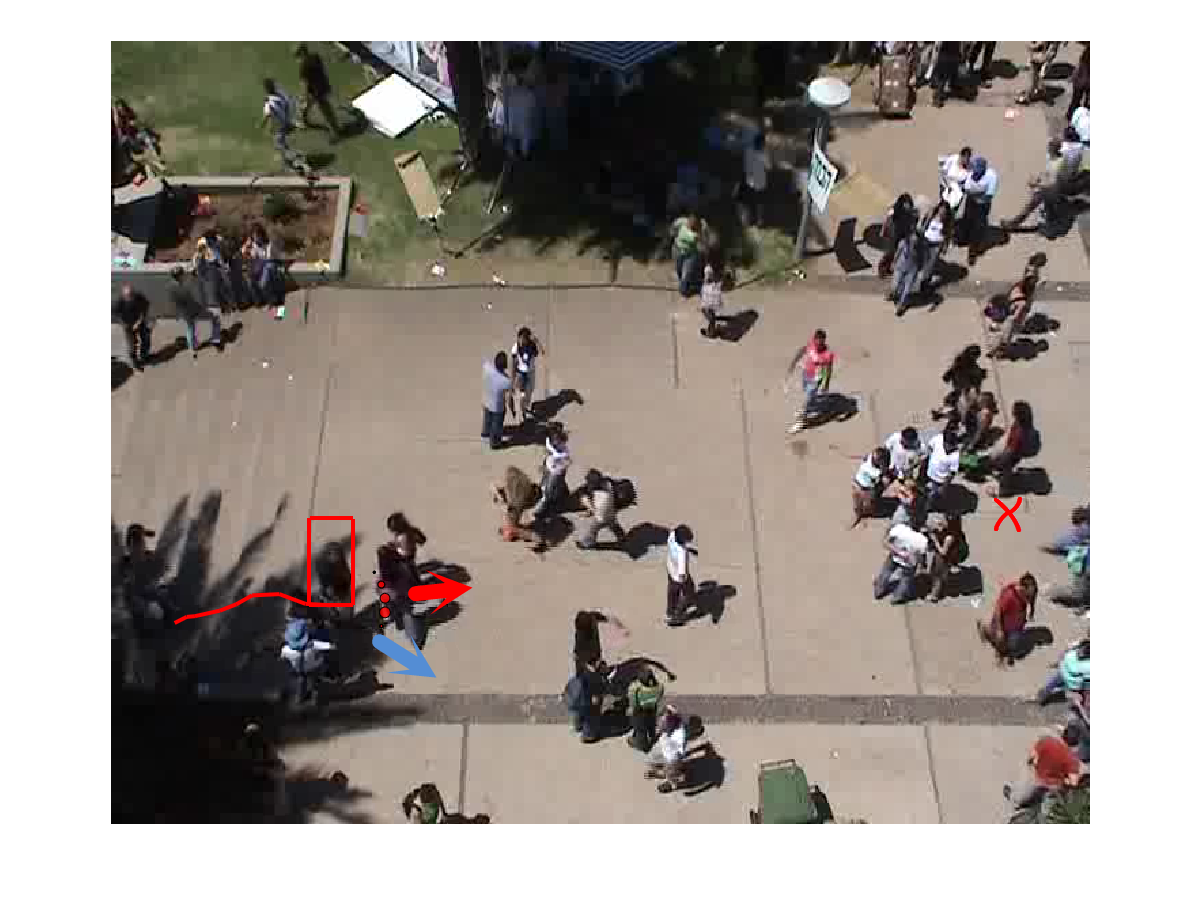}
        \end{subfigure}
        \hspace{-0.3 in}
        \begin{subfigure}[b]{0.32\textwidth}
                \centering
                \includegraphics[width=\textwidth]{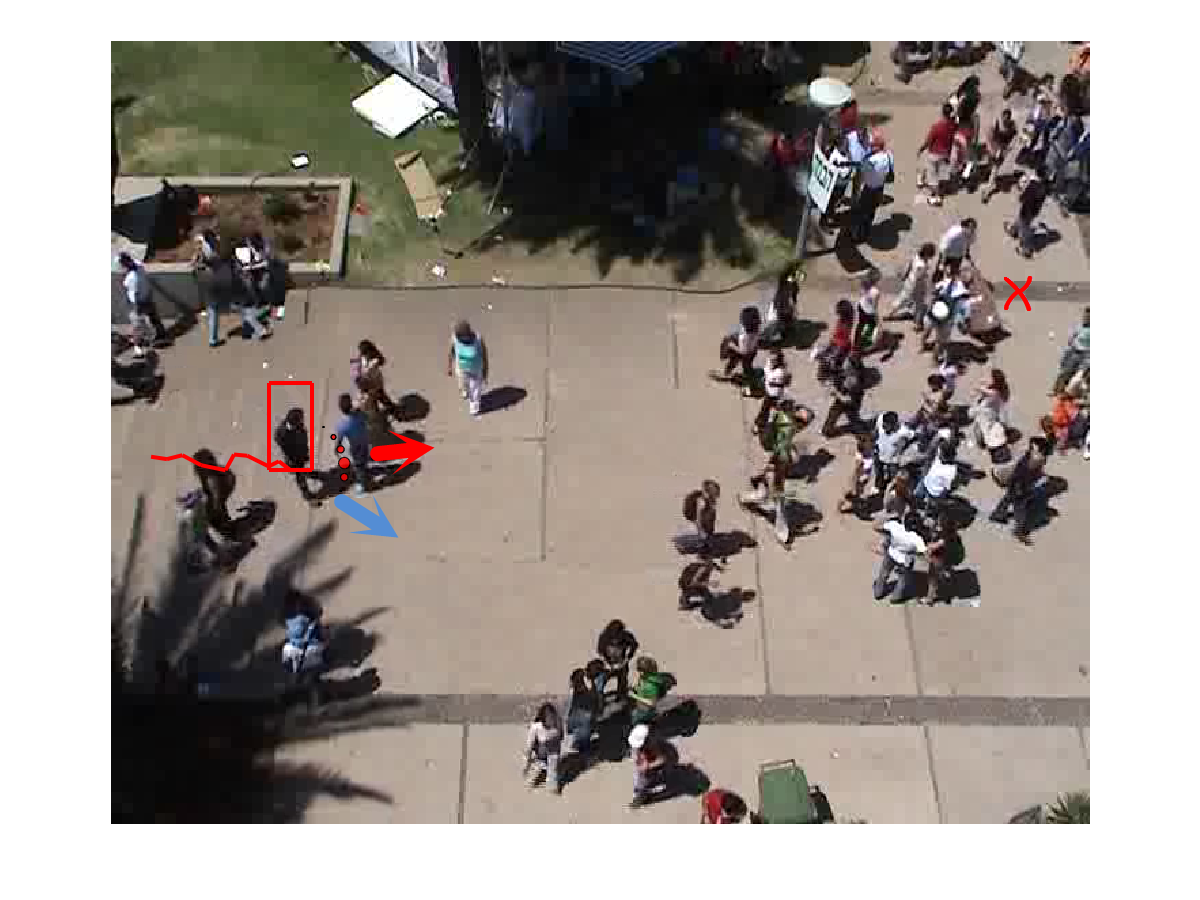}
        \end{subfigure}
        \hspace{-0.3 in}
        \begin{subfigure}[b]{0.32\textwidth}
                \centering
                \includegraphics[width=\textwidth]{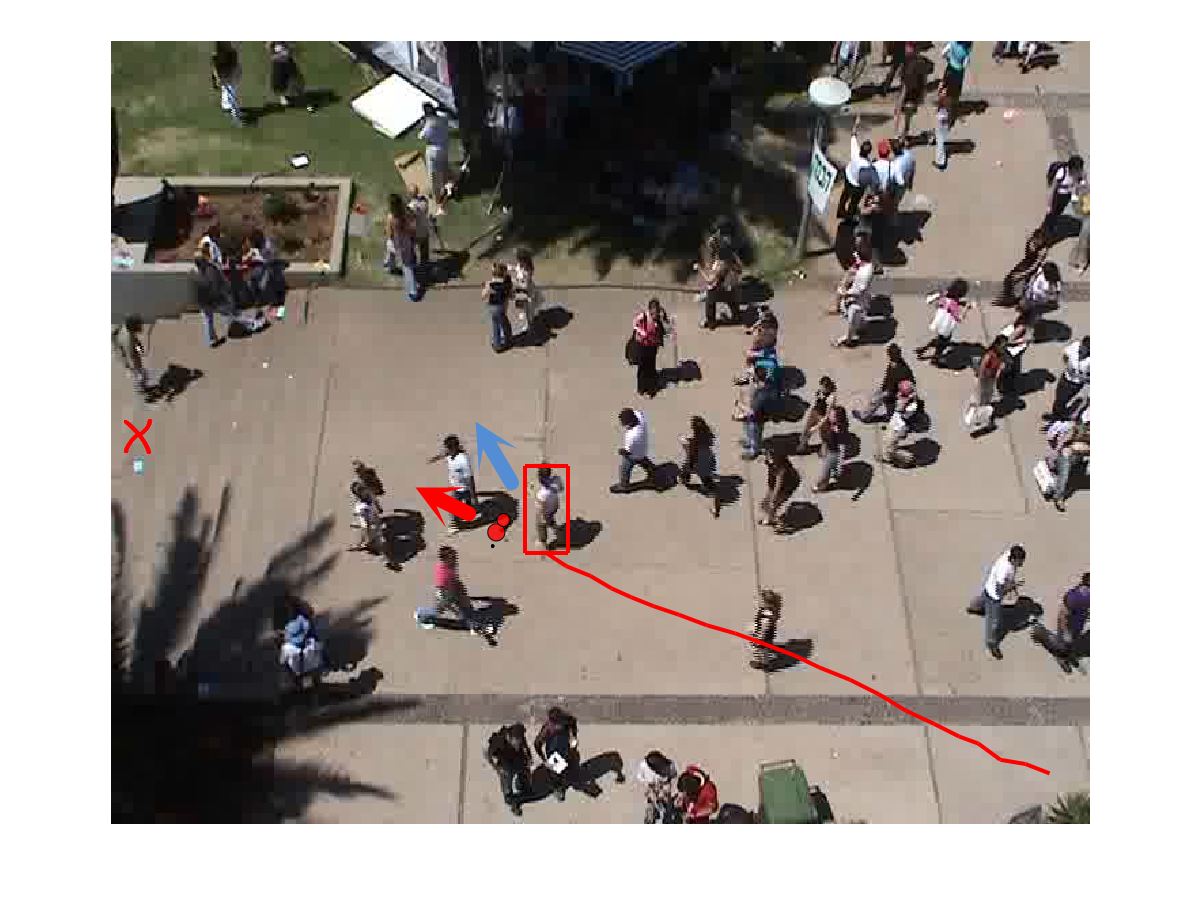}
        \end{subfigure}
        \hspace{-0.3 in}
        \begin{subfigure}[b]{0.32\textwidth}
                \centering
                \includegraphics[width=\textwidth]{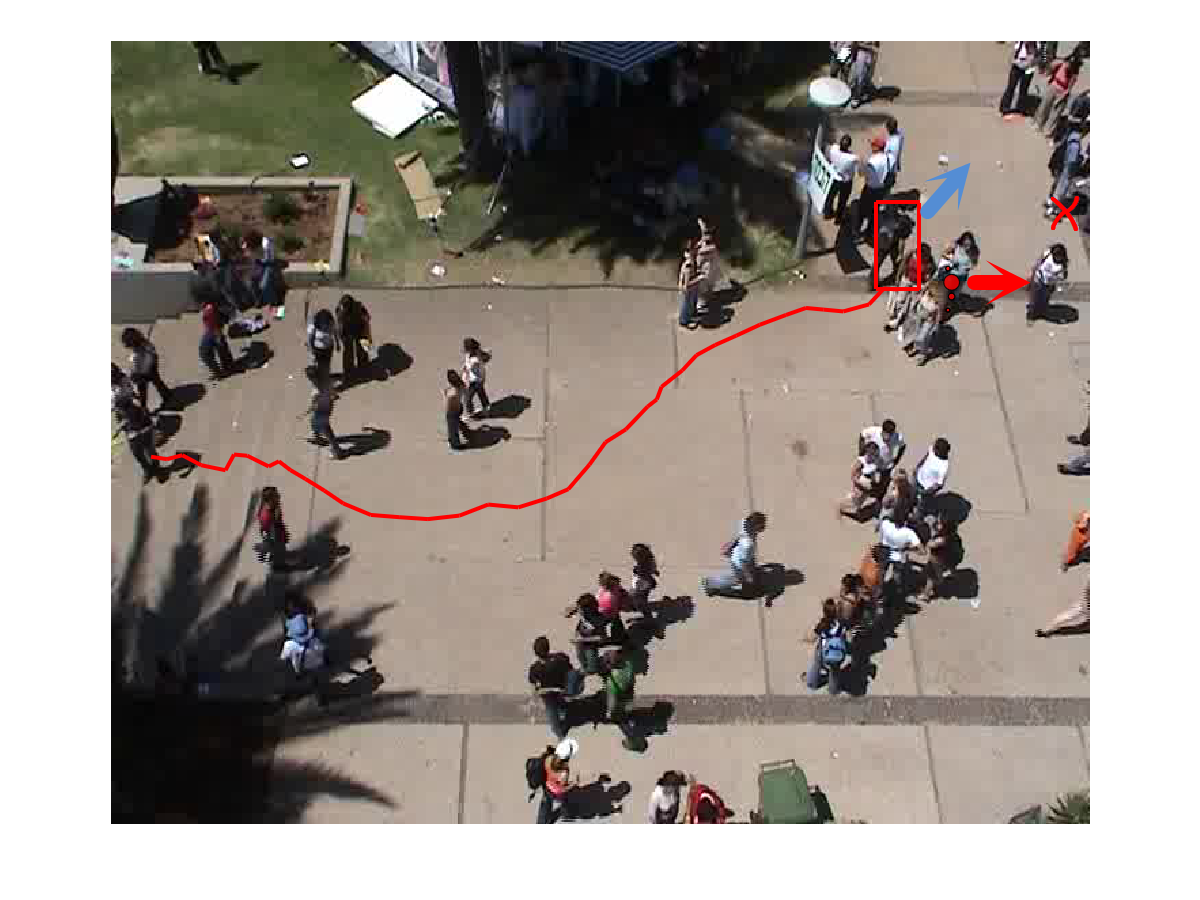}
        \end{subfigure}
        }\\
		\vspace{-0.1in}
        \caption{{\small We show the online-learned desired velocity on \emph{Student}. The blue arrows indicate the current velocities while the red arrows indicate the estimated desired velocities. The red crosses refer to the ground-truth goals of the target persons. The small red circles near the persons refer to particles of desired velocities. (The sizes of the red circles indicate the importances of the particles.) As shown in the examples, the estimated desired velocities remain stable and correct in guiding the tracker towards the destinations even if the person deviates from the correct direction to avoid oncoming pedestrians. }}
        \label{fig:dv}
        \vspace{-0.15in}
\end{figure*}

Using HPF consistently increases STs compared to PF (i.e., RVO+), with an overall ST increase by $\sim$6\%.  In addition, IDSs decrease by 30\% when using HPF, compared to RVO+.
These results suggest that HPF is capable using the longer-term prediction capability of RVO to overcome noise and occlusion.
Figure~\ref{fig:hpf_track_cmp} shows an example of how predictions from previous time-steps
in \mbox{HPF} help overcome occlusions in tracking.
Figure~\ref{fig:track_zara} illustrates four examples showing HPF tracking under frequent interactions and occlusions.
Figure~\ref{fig:track_zara_add} shows additional multi-person tracking results using HPF on $Zara01$, $Zara02$ and $Hotel$. Figure~\ref{fig:ids_zara} and ~\ref{fig:ids_stud} also demonstrate two sets of comparisons, LIN vs HPF and RVO+ vs HPF, respectively. Compared with LIN, HPF takes advantages of RVO's capability in collision-free simulation. As illustrated in Figure~\ref{fig:ids_zara}, the tracker using LIN often switch the identities of pedestrians that walk close to each other, since its underlying motion model does not consider collision avoidance. From Figure~\ref{fig:ids_stud}, the tracker using RVO+ drifts due to the failure of the observation model in highly clustered situation. HPF is able to maintain tracking using higher-order predictions even though there are occlusions. Figure~\ref{fig:dv} show additional examples of how the desired velocity is learned online and helps the trackers. As observed in the figure, the online learned desired velocity is more accurate than the current velocity for the purpose of predicting the final goals.

pHPF (using LIN or RVO+) from \cite{pan2011visual}, which has not been tested in low-fps video, overall performs worse than simple LIN for prediction and tracking.
The reason of its poor performance is two-fold.
First, a new particle is the weighted average of previous particle predictions.  If these predictions are different (but equally valid), then the average prediction will be somewhere in between and may no longer be valid.  Second, the weights are trained offline and do not depend on the observations, which causes \mbox{problems} for particles generated during occlusion.
HPF addresses both these problems by using a multimodal transition distribution in Eq.~\ref{eq:assump}, and using weights that depend on the observations in Eq.~\ref{eqn:lambda}.


Finally, it is worth mentioning that our \mbox{results} (\mbox{LTA+}/ATTR+ in Table~\ref{tab:track_suc}) using online-estimation of \mbox{internal} goals are consistent with \cite{yamaguchi_who_2011}, which uses offline estimation based on labeled goals.
For $N=16$, the total \mbox{STs} on \emph{Zara01}, \emph{Zara02} and \emph{Student} with 2.5fps
are 850/837/840 for LIN/LTA/ATTR in \cite{yamaguchi_who_2011},
and  906/876/882  for LIN/LTA+/ATTR+ reported here.
For $N=24$, the total STs are 394/394/390 for LIN/LTA/ATTR in \cite{yamaguchi_who_2011},
and 432/430/451 for LIN/LTA+/ATTR+ here.
These performance differences among LIN/LTA/ATTR are consistent with those in~\cite{yamaguchi_who_2011} and our results are generally better than these results due to a more accurate observation model.

\section{Conclusions}

We introduce a multiple-person tracking algorithm that improves existing agent-based tracking methods under low frame rates and without scene priors.  We adopt a more \mbox{precise} agent-based motion model, RVO, and integrate it into the particle filter that enables online estimation of the desired velocities. Our PF framework also improves tracking for other agent-based motion models.
Moreover, we derive a higher-order particle filter  to better leverage the longer-term predictive capabilities of the crowd model. 
In experiments,
we demonstrate that our framework is suitable for predicting pedestrians' behaviors, and improves online-tracking in real-world scenes.

\section*{Acknowledgements}
This work was supported by a number of fundings: NSF awards (1000579, 1117127, 1305282), a grant from the Boeing Company, GRFs from the Research Grants Council of Hong Kong (CityU 123212, 110513, 116010), and a SRG from City University of Hong Kong (7002768).

{\small
\bibliographystyle{ieee}
\bibliography{VisionCollection}
}

\end{document}